\documentclass[sigconf]{acmart}
\usepackage[utf8]{inputenc}
\usepackage{hyperref}
\usepackage{xcolor}
\usepackage{graphicx}
\usepackage{array}
\usepackage{caption}
\usepackage{subcaption}
\usepackage{enumitem}
\setlist[itemize]{align=parleft,left=0pt..1em}
\graphicspath{./Figures/}

\acmConference[WWW ’24 Companion]{ACM Web Conference 2024}{May 13--17, 2024}{Singapore}

\title{Rankitect: Ranking Architecture Search Battling World-class Engineers at Meta Scale}
\author{Wei Wen$^*$,\hspace{1em} Kuang-Hung Liu$^{*}$,\hspace{1em} Igor Fedorov,\hspace{1em} Xin Zhang,\hspace{1em} Hang Yin,\hspace{1em} Weiwei Chu}
\thanks{*These authors contributed equally.}
\author{Kaveh Hassani,\hspace{1em} Mengying Sun,\hspace{1em} Jiang Liu,\hspace{1em} Xu Wang,\hspace{1em} Lin Jiang,\hspace{1em} Yuxin Chen}
\author{Buyun Zhang,\hspace{1em} Xi Liu,\hspace{1em} Dehua Cheng,\hspace{1em} Zhengxing Chen,\hspace{1em} Guang Zhao}
\author{Fangqiu Han,\hspace{1em} Jiyan Yang,\hspace{1em} Yuchen Hao,\hspace{1em} Liang Xiong,\hspace{1em} Wen-Yen Chen}
\email{{wewen, khliu, ifedorov, xinzhang5, hyin5, wchu, kavehhassani, mengyingsun, jiangliu, wangxu85, amylinjiang, yuxinc}@meta.com}
\email{{buyunz, xliu1, dehuacheng, czxttkl, gzhao27, fhan, chocjy, haoyc, lxiong, wychen}@meta.com}
\affiliation{%
  \institution{Meta Platforms, Inc., USA}
  \country{}
}
\date{October 2023}
\renewcommand{\shortauthors}{Wen and Liu, et al.}
\begin{document}

\begin{abstract}
Neural Architecture Search (NAS) has demonstrated its efficacy in computer vision and potential for ranking systems.
However, prior work focused on academic problems, which are evaluated at small scale under well-controlled fixed baselines. In industry system, such as ranking system in Meta, it is unclear whether NAS algorithms from the literature can outperform production baselines because of:
% \begin{itemize}[leftmargin=*]
%     \item \textbf{Scale.} Meta's ranking systems serves billions of users
%     \item \textbf{Strong baseline.} The baselines are production models optimized by hundreds to thousands of world-class engineers for years since the rise of deep learning
%     \item \textbf{Dynamic baseline.} During the time it takes NAS to discover a new model, engineers may have established new, stronger baselines
%     \item \textbf{Efficiency.} The search pipeline must yield results quickly in alignment with the productionization life cycle.
% \end{itemize}
\textbf{(1) scale --} Meta's ranking systems serve billions of users,
\textbf{(2) strong baselines --} the baselines are production models optimized by hundreds to thousands of world-class engineers for years since the rise of deep learning,
\textbf{(3) dynamic baselines --} engineers may have established new and stronger baselines during NAS search,
and \textbf{(4) efficiency --} the search pipeline must yield results quickly in alignment with the productionization life cycle.
% Their efficacy is unknown if applied to real-world ranking systems which serve billions of users, such as Facebook users, because
% (1) baselines are very strong. For instance, at Meta, the baselines are production models competed and optimized by hundreds of world-class engineers for years since the rise of deep learning;
% (2) baselines dynamically evolve. NAS requires sufficient time to search, during which, engineers may have established new baselines stronger than production models to battle against NAS models;
% (3) efficiency of NAS is critical, such as, a round of search should accomplish quickly in productionization life cycles.
In this paper, we present \textbf{Rankitect}, a NAS software framework for ranking systems at Meta. 
% We answer this question by conducting \textbf{Rankitect}, a NAS stack for ranking systems at Meta.
Rankitect seeks to build brand new architectures by composing low level building blocks from scratch.
Rankitect implements and improves state-of-the-art (SOTA) NAS methods for comprehensive and fair comparison under the same search space, including sampling-based NAS, one-shot NAS, and Differentiable NAS (DNAS). We evaluate Rankitect by comparing to multiple production ranking models at Meta. We find that Rankitect can discover new models from scratch achieving competitive trade-off between Normalized Entropy loss and FLOPs. When utilizing search space designed by engineers, Rankitect can generate better models than engineers, achieving positive offline evaluation and online A/B test at Meta scale.
% Rankitect targets on building brand new full architectures by composing low level building blocks from scratch.
% Rankitect implements and improves state-of-the-art (SOTA) NAS methods for comprehensive and fair comparison under the same search space, including sampling-based NAS, one-shot NAS and Differentiable NAS (DNAS). We evaluate Rankitect by the most revenue critical ranking model at Meta and generalize it to other model types.
% We find that NAS can discover new models from scratch achieving competitive trade-off between Normalized Entropy and FLOPs when battling world-class engineers. When utilizing search space designed by engineers, Rankitect can generate better models than engineers, achieving positive offline test and online A/B test at Meta scale.

\end{abstract}

\maketitle

\section{Introduction}

Neural Architecture Search (NAS) seeks to automatically generate new architectures for deep learning models.
A seminal work~\cite{zoph2016neural} proved the efficacy of NAS in the era of deep learning to design convolutional networks for computer vision and recurrent cells for natural language processing.
It stimulated a line of academic research, typically in computer vision~\cite{zoph2018learning,wu2019fbnet,tan2019efficientnet} and later in ranking systems~\cite{zhang2023nasrec,gao2021progressive,krishna2021differentiable}. One of the major research areas around NAS has been improving its computational efficiency, including more efficient sampling-based methods~\cite{wen2020neural,real2019regularized,white2021bananas,NEURIPS2019_044a23ca}, one-shot~\cite{pham2018efficient,bender2020can} and DNAS~\cite{xie2018snas,liu2018darts,NEURIPS2022_753d9584,MLSYS2021_c4d41d96}.
However, those methods have never been evaluated at web-scale applications with billions of users for generating brand new ranking models from scratch by composing low-level building blocks.
In this paper, we develop Rankitect, which advances NAS to Meta scale with comprehensive evaluation:
\begin{itemize}[leftmargin=*]
    \item \textbf{Search space}: Our search space includes almost no inductive bias from the production model. The discovered architecture is a mapping between feature embeddings and final classification layer. Building blocks are low-level operations like linear, dot product, etc. Connectivity among blocks is arbitrary. Block dimensions are searchable;
%   \item \textbf{Search space}: search full ranking architecture between original feature embeddings and final classification layer. Building blocks are at the low-level of linear, dot product, and others. Connectivity among blocks is arbitrary. Block dimensions are also searchable;
  \item \textbf{Improved three categories of search algorithms}: sampling-based methods~\cite{wen2020neural,zoph2016neural} under different low cost proxies, one-shot method~\cite{bender2020can} with in-place co-trained baseline model and DNAS~\cite{xie2018snas};
  \item \textbf{Baseline}: the strongest production model at Meta, which is a Click Through Rate (CTR) model optimized by world-class engineers driven by years of business needs. 
\end{itemize}

\noindent This industry level NAS at Rankitect scale imposes significant challenges and we advance current SOTA methods to succeed:

\begin{itemize}[leftmargin=*]
  \item \textbf{Low cost evaluation analysis}: for sampling-based methods, training a sampled model for long term is unrealistic, e.g. $>50$ billion examples at Meta scale. When a small amount of data is used, we compare the ranking quality between weight-sharing proxy~\cite{bender2018understanding} and early-stop proxy, and find weight-sharing proxy is superior when the data amount is small but will be surpassed by simple early-stop proxy when more data is used; based on our decision framework, early stop proxy is preferred in practice when requiring a higher ranking quality and more efficient engineering and system optimization.
  \item \textbf{Noise reduction by an in-place baseline in one-shot method}: data distribution of real-world ranking systems shifts dramatically, which makes the absolute metric -- normalized entropy (NE) loss~\cite{he2014practical} -- noisy over the training course. To cancel the noise introduced by data shifting, we co-train a baseline model in parallel with the weight sharing supernet in one-shot method~\cite{bender2020can}, and use the relative improvement of a model/subnet sampled from the supernet versus the baseline as the reward of the Reinforcement Learning (RL) agent;
  \item \textbf{A joint on-policy/off-policy RL method to improve sample efficiency}: RL based one-shot NAS suffers from costly reward evaluation and extreme sample inefficiency. To address both challenges, we propose a joint on-policy and off-policy policy gradient method that uses experience replay to amortize the computational cost of reward evaluation and improve sampling efficiency by decoupling RL policy updates from that of weight sharing supernet.
  \item \textbf{DNAS}: We implement differentiable NAS (DNAS) to optimize decision variables in conjunction with supernet weights. The DNAS approach exhibits increased efficiency compared to sampling-based methods and improves convergence characteristics compared to one-shot method using RL.
  \item \textbf{System optimization}: Optimizing NAS within expensive search spaces presents inherent computational challenges in production system. To overcome this challenge, we developed tailored techniques such as: (1) activation checkpointing; (2) employing system-wide vectorized operations; (3) utilizing transfer learning to facilitate supernet weight sharing; (4) Round-Robin process groups to mitigate AllReduce overhead; (5) dynamic execution of partial operators in PyTorch graphs. These strategies collectively enhance the efficiency and effectiveness of Rankitect search.
    % \item \textbf{System optimization}: Optimizing NAS within expansive search spaces presents inherent challenges in managing memory and improving training efficiency. During supernet search, dynamic computation graphs for search candidates, which are not well-suited for conventional optimizations, are encountered. To streamline this process, we have explored and implemented several techniques, including activation checkpointing for reduced memory usage, employing vectorized operations to expedite computations, leveraging GPU and CPU trace analysis to identify and optimize inefficiencies like deepcopy operations, utilizing transfer learning to facilitate weight sharing between the supernet and standalone subnets for efficient one-shot training, and implementing Round-Robin process groups to mitigate AllReduce overhead\cite{li2020pytorch}. These strategies collectively enhance the efficiency and effectiveness of neural architecture search, especially in the context of large search spaces and advanced search algorithms.
%   \item \textbf{FLOPs constraint???}: \textcolor{red}{Wei}. 
\end{itemize}

\noindent With improved NAS methods, we find that
\begin{itemize}[leftmargin=*]
  \item Rankitect can generate brand new ranking models from scratch and outperform production models. Three methods in Rankitect all generated models along the NE-FLOPs Pareto front, but one-shot and DNAS are about $145\times$ efficient in terms of GPU hours. While one-shot and DNAS are on-par in terms of model quality and search efficiency, we find one-shot method can be implemented more easily with higher memory efficiency.
  \item During the dynamic iteration of the production model, in-house world-class engineers were also able to invent models beating the baseline. When battling human models,  Rankitect models can cover different regions along the Pareto frontier of NE-FLOPs. Moreover, by utilizing the human designed model as a search space, Rankitect produces models with NE-FLOPs trade-off better than engineers achieving positive offline evaluation and online A/B test at Meta scale.
\end{itemize}

% comprehensive comparison of methods in the same benchmark at industry level

\section{Related Work}
Neural Architecture Search (NAS) is a generic approach to design deep learning models in various applications, such as, computer vision~\cite{pham2018efficient,bender2020can,wen2020neural,real2019regularized,wu2019fbnet,tan2019efficientnet,zoph2018learning,zoph2016neural,9157751}, natural language processing~\cite{zoph2016neural,klyuchnikov2022bench,so2019evolved} and speech~\cite{mehrotra2020bench,abdelfattah2021zero}.
Its adoption to ranking systems is also ramping up~\cite{zhang2023nasrec,krishna2021differentiable,gao2021progressive,song2020towards,zhang2023distdnas,cao2023farthest}, however those work evaluated NAS with small public datasets using small search space.
For example, the DNAS (for ads CTR prediction~\cite{krishna2021differentiable}) and PROFIT~\cite{gao2021progressive} only search a portion of ranking models with a limited number of choice options. NASRec~\cite{zhang2023nasrec} scales the search space to a larger extent but the number of choices to select building blocks is just $7$.
Unlike those work, Rankitect evaluates web-scale NAS at Meta scale with a much larger dataset ($>50$ billions of examples) and a much larger search space ($\geq 28$ choices of building blocks).

NAS has also been applied to industry level ranking systems~\cite{liu2020autofis,li2023hyperscale,zhaok2021autoemb,anil2022factory}, however, the search space is relatively trivial: AutoFIS~\cite{liu2020autofis} uses DNAS like gates to select predefined feature interactions for Huawei App Store ranking systems; ByteDance proposed AutoEmb~\cite{zhaok2021autoemb} to learn embedding dimensions, and Google's H$_{2}$O-NAS~\cite{li2023hyperscale} is adopted to learn MLP sizes and embedding dimensions. None of them has studied the aggressive search space like Rankitect, which searches building block per layer, connections among blocks and dimensions per block, all at once..

Moreover, only one NAS algorithm was adopted in previous work,
such as sampling-based methods in AutoCTR~\cite{song2020towards} and NASRec~\cite{zhang2023nasrec}, DNAS in AutoFIS and one-shot method in H$_{2}$O-NAS. This imposes challenges to understand the scalability and effectiveness of different NAS algorithms because the setups are different across different works/companies. Rankitect evaluates three categories of NAS algorithms (sampling-based methods, DNAS and one shot method) using the same dataset, search space, baseline and engineering constraint.
This provides valuable insights to select appropriate NAS methods based on real-world conditions.

\section{Method}
In this section, we introduce the search space and three NAS algorithms we developed.

\subsection{Search Space and Supernet  Design}
\begin{figure*}
  \centering
  \includegraphics[width=\textwidth]{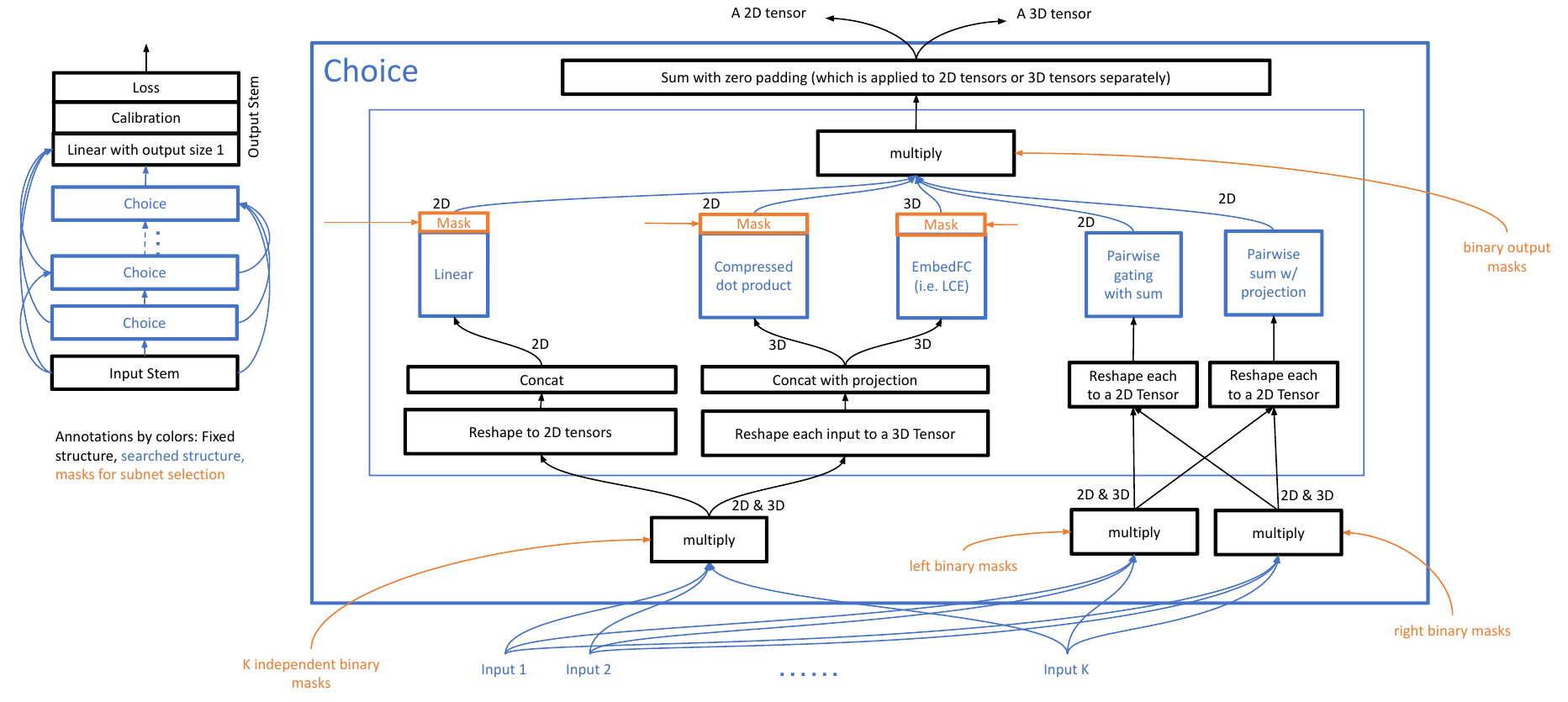}
  \caption{Search space in sampling-based methods and supernet in one-shot method and DNAS}
  \label{fig:supernet}
\end{figure*}

Search space is defined as the set of all possible models. In Rankitect, we bag all possible models into a single weight sharing supernet model~\cite{bender2018understanding} as detailed in Figure~\ref{fig:supernet}, and a model in the search space is produced by masking out connections \& building blocks and selecting specific dimensions for building blocks. As a selected model is materialized as a sub-architecture of the supernet, we interchangeably use ``subnet'' and ``model''.

% As discussed in the Introduction, the supernet itself is the search space. The first step of WS-NAS is to design a supernet (i.e. a search space) to cover all interested models. When we put more possible architectures into the search space, the size of search space becomes larger, the NAS becomes more automated without requiring human expert knowledge and it is more flexible to search more possible architectures; however, the speed of a single search becomes slower since there are more architectures to train, and the optimization becomes harder since finding a good model in a very large search space is like looking for a needle in a haystack. Vice versa. When we shrink the search space, human expert knowledge is required to ensure that the small search space includes enough good models; therefore, we sacrifice the automation and flexibility of NAS, while the merits are faster search speed and easier optimization. If Deep Learning shifts machine learning from feature engineering to architecture engineering, NAS is likely shifting us to search space engineering.
Rankitect searches the end to end architecture between raw inputs (a dense feature 2D vector and a 3D embeddings concatenated from sparse/category embeddings and content embeddings) and the final logit used for CTR prediction.
For simplicity, a 3D or 2D tensor is denoted as $X_{3d} \in \mathbb{R}^{B\times N \times D}$ or $X_{2d} \in \mathbb{R}^{B\times S}$, respectively. $B$ is batch size, $N$ is the number of embeddings, $D$ is embedding dimension, and $S$ is the size of a 2D vector.
In Rankitect, all 3D tensors have the same embedding dimension $D$.

The supernet is constructed by stacking a cascade of “Choice” modules which can be fully densely connected to previous choices and raw inputs.
All Choice modules have the same architecture which consists of three parts:
\begin{enumerate}[leftmargin=*]
    \item \textbf{Input adaptors}, which adapt inputs for valid computation of each building “block”. We have three types of adaptors as shown in Figure~\ref{fig:supernet}: (1) reshape each input to a 2D tensor and concatenate together before the linear block; (2) concatenate all inputs to a single 3D tensor along the middle axis (when an input is $X_{2d} \in \mathbb{R}^{B\times S}$, it is expanded to $X_{3d} \in \mathbb{R}^{B\times 1 \times S}$ and then linearly projected to $X_{3d} \in \mathbb{R}^{B\times 1 \times D}$ for concatenation with other 3D tensors). This adaptor is used before compressed dot product and EmbedFC blocks; (3) reshapes each input to a 2D tensor before pairwise gating and sum blocks.
    \item \textbf{Building blocks}, which are layer types we want to search. Rankitect generates models from the elementary low level building blocks below:
    
    \begin{itemize}[leftmargin=*]
    \item \textbf{linear} or fully connected (FC) layer
    \item \textbf{EmbedFC} is a linear projection applied to the middle axis of input $X_{3d} \in \mathbb{R}^{B\times N \times D}$ and produces $X_{3d} \in \mathbb{R}^{B\times M \times D}$, therefore, it is a linear with weight matrix $W \in \mathbb{R}^{M \times N}$;
    \item \textbf{compressed dot product} is simply pairwise dot products among compressed embeddings and raw embeddings as
    \begin{equation}
    flatten \left(bmm \left(X_{3d}, EmbedFC(X_{3d})^T \right) \right),
    \end{equation}
    where $bmm(\cdot)$ is batch matrix-matrix product~\footnote{https://pytorch.org/docs/stable/generated/torch.bmm.html};
    \item \textbf{pairwise gating} block has $K$ inputs $X_{2d}^{(i)} \in \mathbb{R}^{B\times S_i}$ where $i \in {1...K}$, it computes as 
        \begin{equation} \label{eq:pairwise}
        \sum_{i=1}^{K} \sum_{j=1}^{K} interaction\left( FC_{ij}\left( X_{2d}^{(i)} \right), X_{2d}^{(j)} \right),
        \end{equation}
        where $FC_{ij}(\cdot)$ projects $X_{2d}^{(i)}$ from dimension $S_i$ to $S_j$ and $interaction(x, y) = sigmoid(x) \circ y$ known as Hadamard product. $\sum_{i=1}^{K} \sum_{j=1}^{K} (\cdot)$ will zero pad smaller tensors to maximum size of $\{ S_i | i \in {1...K} \} $ before sum;
    \item \textbf{pairwise sum} block is very similar to pairwise gating block and the only difference is that $interaction(x, y) = x + y$.
    
    \end{itemize}
    
       Note that firstly layer normalization and then activation functions are appended after each block.

    \item \textbf{Output aggregators}, which reduce outputs from all building blocks. In Rankitect, all 2D block outputs are zero-padded to the same shape and then sum together to form a single 2D tensor. As only one block (EmbedFC) produces a 3D tensor, the 3D tensor is outputted alone from the choice module.
\end{enumerate}
In the supernet as colored by \textcolor{orange}{orange} in Figure~\ref{fig:supernet}, Rankitect samples masks to sample a subnet/model from the search space. A subnet is sampled by selecting a building block within each choice module, choice connections, and block dimensions:
\begin{itemize}[leftmargin=*]
    \item \textbf{building block sampling:}  Rankitect selects one block within each choice module, which is accomplished by sampling an one-hot ``binary output masks''.
    \item \textbf{choice connection sampling:} when linear, compressed dot product or EmbedFC block is selected, ``K independent binary masks'' (multi-hot masks) are sampled to select a portion as inputs; when pairwise gating or sum block is selected, an one-hot ``left binary masks'' and one-hot ``right binary masks'' are sampled to enable only one pair of ``$interaction(\cdot)$'' in Eq.~(\ref{eq:pairwise}).
    \item \textbf{block dimension sampling:} when a block is linear, compressed dot product or EmbedFC, output dimension of the linear projection inside the block will be sampled. This is achieved by multiplying linear outputs by a vector of binary masks. A mask vector is constrained to values with leading ones followed by zeros (e.g. $[1,1,1,0,0]$). The length of leading ones equals an selected output dimension.
\end{itemize}
Note that, Rankitect can enable the whole supernet by setting all masks values to ones.
With our supernet design, Rankitect flexibly supports a variety of NAS algorithms:
\begin{itemize}[leftmargin=*]
\item sampling-based algorithms: any black-box sampling algorithms (random, reinforcement learning~\cite{zoph2016neural}, Bayesian optimizer~\cite{white2021bananas} and neural predictor~\cite{wen2020neural}) can be used to sample masks (in orange) to sample a subnet/model;
\item one-shot method: an in-place RL agent is co-trained with the supernet, with the agent sampling masks;
\item DNAS: masks are replaced by Gumbel-Softmax
~\cite{jang2016categorical} learned by architecture parameters.
\end{itemize}

% Finally, following lists are different strategies we considered when designing a supernet (from high risk to low risk but from more headroom to less headroom)
% \begin{enumerate}
%     \item A full supernet, which includes all building blocks in any choices and all choices are densely connected.
%     \item A supernet grown from our production model by aggressively adding building blocks and connections. This is the approach we currently adopted.
%     \item A supernet customized with human prior for specific model architecture exploration, such as a DHEN-like supernet for large capacity models, a supernet for transformer exploration, etc.
% \end{enumerate}

% search space size: O(???)
\subsection{NAS Algorithms}
At Meta, we use Normalized Entropy (NE)~\cite{he2014practical} loss to evaluate the prediction performance of ranking systems.
A smaller NE presents a better ranking model. We use ``NE gain'' over a baseline to measure improvement -- a negative ``NE gain'' implies a improved prediction.
In the following, we introduce how each NAS algorithm works in Rankitect and what improvement we have proposed.

\subsubsection{Sampling-based method}\hfill\vspace{0.5em}

Based on study in computer vision on NASBench-101~\cite{ying2019bench,wen2020neural}, Neural Predictor~\cite{wen2020neural} is a more efficient sampling-based method than Regularized Evolution~\cite{real2019regularized} and Reinforcement Learning, we select Neural Predictor given it also has traits of full parallelism and friendly hyper-parameter tuning. Unlike original method which used a random sampler to pick top models by the predictor, we use a RL sampler to do so for faster convergence.

As discussed, it is unrealistic to do a full job training at Meta dataset scale after a subnet/model is sampled.
We compared two types of low cost NE proxies:
\begin{itemize}
    \item \textbf{weight-sharing proxy} which has two stages: \textbf{supernet pretraining} and \textbf{subnet finetuning}.
    In \textbf{supernet pretraining}, the whole supernet is first trained for $10\%$ data (by setting all masks to ones), followed by enabling subnet sampling with a proability of $0.75$ at each mini-batch. That is, in the later $90\%$ supernet pretraining, at each mini-batch, the supernet has $0.25$ probability to train all architectures and $0.75$ probability to only train a sub-architecture (subnet) by uniform random sampling of masks (in orange in Figure~\ref{fig:supernet}). In \textbf{subnet finetuning}, the Neural Predictor will first sample and fix masks to select a subnet, and then the specific subnet (with weights shared/transferred from corresponding parameters from supernet) is fine-tuned for a small amount of data to get the final weight-sharing proxy. Note that fine-tuned weights in a sampled subnet will not be updated to the supernet.
    \item \textbf{early-stop proxy}, which simply trains a standalone subnet/model from scratch for little data. For this proxy, the supernet is never instantiated or trained; only the sub-architecture of a sampled model is realized as a PyTorch model for more efficient training.
\end{itemize}

Note that for both NE proxies of a subnet, window NE averaged over the last $25\%$ training/fine-tuning data is used because it  better correlates with long term NE which is more important for CTR models in real-world ranking systems.

\subsubsection{One-shot NAS with Reinforcement Learning}\hfill\vspace{0.5em}
\label{sec:one_shot_rl}

\begin{figure}
\centering
\includegraphics[width=0.4\textwidth]{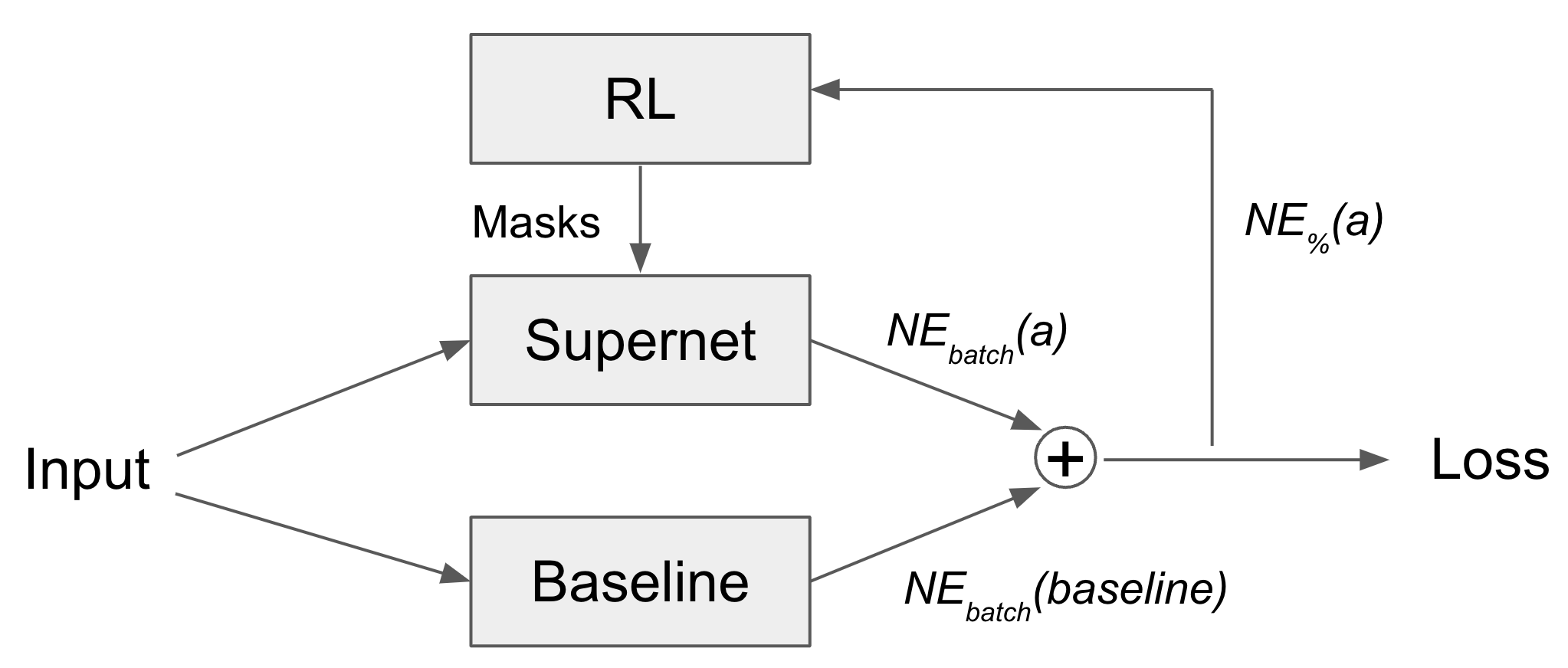}
\caption{In-place training a baseline model with RL sampling subnets from the supernet.}
\label{fig:rl_baseline_reward}
\end{figure}

Motivated by the TuNAS one-shot method~\cite{9157751}, we co-train an in-place RL agent with the supernet simultaneously illustrated in Figure~\ref{fig:rl_baseline_reward}. At each mini-batch, the RL agent samples masks to select a subnet and the mini-batch NE is used as reward to update the RL agent. More specific, we treat all the decisions (i.e., mask selections in the supernet) as independent multinomial \textbf{r}andom \textbf{v}ariables (r.v.). Let $\mathbf{\theta}$ be a vector (collection) of parameters that define all the decision's multinomial r.v.. We use REINFORCE~\cite{10.1007/BF00992696} to update the sampling distribution $P_{\mathbf{\theta}}$ with policy gradient: 
\begin{equation}
    \nabla_{\mathbf{\theta}} \sum_i \log P_{\mathbf{\theta}}(a_{i}) (R_{i} - b),
    \label{eq:pg_orig}
\end{equation}
where $a_{i}$ is a sampled subnet and $R_{i}$ is the reward (e.g. NE and FLOPs) of $a_{i}$.
A baseline function $b$ is used to reduce the variance of policy gradient and in our work we define $b$ to be the average over all $R_{i}$. In the following, we discuss technical challenges of this TuNAS-like one-shot method in Rankitect and our proposals to overcome them.
We use ``RL method'' and ``one-shot method'' interchangeably in this paper. 

\paragraph{\textbf{Noisy reward and variance reduction}}

When applying RL to NAS for ranking problem, a main component of $R_{i}$ is the mini-batch NE, $NE_{batch}(a_{i})$. However, $NE_{batch}(a_{i})$ in production environment is: (1) extremely noisy (i.e., high variance) and (2) non-stationary caused by data distribution shift. To overcome these challenges, we borrow ideas from the control variate method where we subtract a correlated baseline reward, denoted as $NE_{batch}(baseline)$, from $NE_{batch}(a_{i})$. More specifically, we propose the following:
\begin{enumerate}[leftmargin=*]
    \item Calculate $NE_{batch}(baseline)$ from an in-place baseline model co-trained in parallel with the supernet.
    \item Replace $NE_{batch}(a_{i})$ used in $R_{i}$ with $NE_\%(a_{i})$:
    \begin{equation}
        NE_\%(a_{i})=\frac{NE_{batch}(a_{i}) - NE_{batch}(baseline)}{NE_{batch}(baseline)}
        \label{eq:ne_gain}
    \end{equation}
\end{enumerate}
Figure~\ref{fig:rl_baseline_reward} illustrates how the proposed $NE_{batch}(baseline)$ and $NE_\%(a_{i})$ integrate with the weight sharing supernet training. A detailed ablation study for this part is included in Appendix~A (see Figure~\ref{fig:neBatch_vs_neGain_reward} and \ref{fig:neBatch_vs_neGain_ne})

\paragraph{\textbf{Sample scarcity and on/off-policy RL}}

REINFORCE is an on-policy method where it can only learn through sampling from the current $P_{\mathbf{\theta}}$. This limitation introduces inefficiency in NAS for industry-scale ranking problems, because of:
\begin{enumerate}[leftmargin=*]
    \item costly reward evaluation: evaluating $R_{i}$ for an industry-scale model architecture is computationally expensive such that the number of samples is prohibited; and
    \item extreme sampling inefficiency: RL policy's updating schedule is slower than that of the supernet update (i.e., Eq.~\eqref{eq:pg_orig} requires a batch of model evaluations where each needs a mini-batch of subnet/supernet training). Consider an example of using batch size 100 for Eq.~\eqref{eq:pg_orig}, then a supernet trained over one million steps (typical budget) only generates $10,000$ RL policy updates which is a challenge for on-policy RL algorithms.
\end{enumerate}
To address these challenges we adopted off-policy RL methods reusing past samples $(a_{i}, R_{i})$ for policy updates with experience replay \cite{DBLP:journals/corr/MnihKSGAWR13}. This allows us to amortize the computational cost of reward evaluation and improves sample efficiency by decoupling policy updates from supernet updates. We followed a standard formulation of the off-policy policy gradient (PG) based on importance sampling (IS) \cite{10.5555/3020847.3020905}. A main challenge of off-policy implementation is controlling for the possible high variance (infinite variance in some cases) of the IS weights (a ratio of two probability functions). In this work, we adopted two techniques to mitigate the potential problem of high IS weights variance: (1) IS weights clipping; (2) weighted importance sampling (WIS) \cite{10.5555/3020847.3020905, 10.5555/3454287.3454448, JMLR:v21:20-124}. In summary, we adopted the following policy gradient with off-policy IS:
% \begin{equation}
%     \nabla_{\mathbf{\theta}} \sum_i \frac{w_{i}}{\sum w_{i}} \log P_{\mathbf{\theta}}(a'_{i}) (R'_{i} - b'), \;\;(\mathbf{\theta}'_{i}, a'_{i}, R'_{i})\sim B,
%     \label{eq:pg_is}
% \end{equation}
\begin{align}
    \nabla_{\mathbf{\theta}} \sum_i \frac{w_{i}}{\sum_j w_{j}}  & \log P_{\mathbf{\theta}}(a'_{i}) (R'_{i} - b'), \;\;(\mathbf{\theta}'_{i}, a'_{i}, R'_{i})\sim B,
    \label{eq:pg_is} \\
    w_{i} &= \min\left[\frac{P_{\mathbf{\theta}}(a'_{i})}{P_{\mathbf{\theta}'_{i}}(a'_{i}) + \epsilon}, 1e4 \right].
\end{align}
where $\epsilon$ is a small number to avoid divide-by-zero, $w_{i}$ is the WIS weights, $b'$ is the average over all $R'_{i}$, and $(\mathbf{\theta}'_{i}, a'_{i}, R'_{i})$ is a random sample drawn from a replay buffer $B$ that stores past samples. Note that we also store the multinomial r.v. parameter $\mathbf{\theta}'$ that are used to sample $a'_{i}$.
% \begin{equation}
%     \min\left[\frac{P_{\mathbf{\theta}}(a'_{i})}{P_{\mathbf{\theta}'_{i}}(a'_{i}) + \epsilon}, 1e4 \right].
% \end{equation}
Since RL policy is trained concurrently with the supernet, recent $(a'_{i}, R'_{i})$ samples are more relevant/accurate than older samples. Based on this observation, we propose a joint on-policy and off-policy PG (denoted as on/off-policy PG) where we first perform on-policy update using Eq.~\eqref{eq:pg_orig} (and store new batch of $(\mathbf{\theta}'_{i}, a'_{i}, R'_{i})$ to $B$) followed by $K$ off-policy updates using Eq.~\eqref{eq:pg_is}. We also experimented with prioritized experience replay \cite{DBLP:journals/corr/SchaulQAS15} but found our proposal that guarantee on-policy update at each step to perform better in practice . With the proposed on/off-policy method, we can achieve a $K\times$ increase in the number of policy updates compared to on-policy REINFORCE. A detailed ablation study for this part is included in Appendix (see Figure~\ref{fig:pg_vs_offpg_reward}).

\paragraph{\textbf{NE search with FLOPs constraint}}

To promote the discovering of models with good NE and FLOPs trade-off, we augment reward function with an additive cost term:
\begin{equation}
    R_{i} = NE_\%(a_{i}) + \alpha \cdot Cost(a_{i}),
\end{equation}
where $\alpha$ is a regularization weight that balances the trade-off between NE and FLOPs.
% which is typically measured by training/deployment latency. In practice, measuring latency is difficult as it depends on the target software-hardware stack and in this work we approximate latency with model size/FLOPs. 
For $Cost(a_{i})$, we experimented with L1-penalty $\|FLOPs(a_{i})-c\|$ and ReLU-penalty $\max(FLOPs(a_{i})-c, \; 0)$, where $c$ is the target FLOPs. From our experiment, we observe that using ReLU-penalty often leads to a model with smaller FLOPs and worse NE compared to using using L1-penalty. Our hypothesis is that the L1-penalty can better guide RL search during initial exploration towards the targeted FLOPs (penalize both above and below target FLOPs) and later lead to the convergence of a better local optimum model.

\subsubsection{DNAS}\hfill\vspace{0.5em}

We implemented DNAS \cite{liu2018darts,cai2018proxylessnas,Dong_2019_CVPR} as an alternative to one-shot method for comprehensive study. DNAS directly optimizes the pair $(W, \theta)$:
\begin{align}
\min_{W, \theta: FLOPs(\gamma(\theta)) \leq c} E_{a, d} \left[CE(W,a,d) \right] \label{eq:dnas_objective}
\end{align}
where $\gamma(\theta)$ denotes the most likely subnet given $\theta$, $d$ is a batch of data, and $CE$ is the cross-entropy loss. The fundamental difference between RL and DNAS is in how we deal with the non-differentiable expectation over $a$ in \eqref{eq:dnas_objective}. Each of the decisions of $a$ is a categorical r.v., such that differentiation of \eqref{eq:dnas_objective} is not possible. In DNAS, the categorical r.v.'s are replaced by Gumbel-Softmax r.v.'s, which lend themselves to the reparameterization trick for differentiation \cite{kingma2022autoencoding,NEURIPS2019_044a23ca,krishna2021differentiable}. The expectation in \eqref{eq:dnas_objective} is approximated using Monte-Carlo sampling, with each trainer and each minibatch element using a separate realization of $a$.
% for a total of $TB$ realizations. 
In order to enforce the constraint in \eqref{eq:dnas_objective}, we follow \cite{NEURIPS2022_753d9584} and transform \eqref{eq:dnas_objective} into a regularized objective with a regularizer whose minimizer is guaranteed to satisfy the constraint:
\begin{align}
    R = E_a \left[ \left | FLOPs(a) - c \right | \right]
\end{align}

% To summarize this section, we present a comparison of the three discussed methods---sampling-based, DNAS, and RL---in Table~\ref{table:search_comparison}.
% \begin{table}[ht]
% \caption{Comparison of different search algorithms in Rankitect}
%  \centering
% \begin{tabular}{|m{5em}|m{5em}|m{6em}|m{6em}|} 
%  \hline
%  \textbf{Method} & Sampling-based & DNAS & RL \\
%  \hline
%  \textbf{Search speed} & slow & fast & fast \\
%  \hline
% \textbf{Memory \hspace{0.05em} efficiency} & High & Low (need to train supernet) & Low (need to train supernet)\\
%  \hline
%  \textbf{Search space} & Architecture and Non-architecture & Architecture only & Architecture only \\
%  \hline
%  \textbf{{\footnotesize Differentiable}} & No & Yes & No \\
%  \hline
%  \textbf{Stability} & High & Need to carefully tune hyperparameters & RL is robust to noise \\
%  \hline
%  \textbf{{\footnotesize Regularization (e.g. FLOPs)}} & Easy & Need to make them differentiable & Easy \\
%  \hline
% \end{tabular}
% \label{table:search_comparison}
% \end{table}

\section{Experiments}

In this section, we provide experimental results for end-to-end Rankitect search and compare the discovered model to strong production baselines created by ML experts.
Three categories of algorithms are also compared.

% https://fb.workplace.com/notes/1200908290779460/
% https://fb.workplace.com/notes/298585006056840/

% https://fb.workplace.com/notes/146411041508295/
% https://fb.workplace.com/notes/163299336401755/

\subsection{Experiment Setup}
We use industrial ranking systems' datasets at Meta for all experiments.
Input features include floating-point dense features, content embeddings, and sparse embeddings from categorical lookup tables. All embeddings are concatenated to a single 3D tensor.
Click Through Rate (CTR) and Conversion Rate (CVR) models are evaluated, which are binary classifiers.
Binary cross entropy loss is used for optimization, and we report Normalized Entropy~\cite{he2014practical} (NE). NE is simply normalized cross entropy which is less noisy over data distributions.
A typical ranking model is trained by more than $50$ billion examples with Adam optimizer.
All Rankitect searches are performed in GPU (up to 128) training cluster. During supernet search, PyTorch dynamic computation graphs are not well-suited for conventional optimizations. To streamline this process, we have explored and implemented several techniques, including activation check-pointing for reduced memory, employing vectorized operations to expedite computations, 
%leveraging GPU and CPU trace analysis to identify and optimize inefficiencies like deepcopy operations, 
utilizing partial parameter transfer from the supernet to standalone subnets to efficiently obtain weight-sharing proxy, and implementing Round-Robin process groups to mitigate All-Reduce overhead~\cite{li2020pytorch}.
%These strategies collectively enhance the efficiency and effectiveness of neural architecture search, especially in the context of large search spaces and advanced search algorithms.

\subsection{Full Architecture Search from Scratch Battling World-class Engineers}
\label{sec:full_space_exp}

\subsubsection{Supernet Design}\hfill\vspace{0.5em}

\label{sec:supernet_design}
To study the scalability of Rankitect, this section performs NAS in a search space as large as possible, using the largest and strongest CTR model baseline (dubbed as ``CTR app 1'') at Meta.
If we naively use all blocks within each choice and fully connect them, the supernet size explodes easily.
For example, a single full choice already has $>8.5$ billion dense parameters, major contributors of which are pairwise gating and sum blocks.
To reduce supernet size, we propose two optimizations:
\begin{itemize}
    \item Sharing pairwise linear projections $FC_{ij}(\cdot)$ in Eq.~(\ref{eq:pairwise}): when the same pairwise interaction (gating or sum) appears in different choices, the interaction is computed once and shared;
    \item bottleneck layers in pairwise linear projections: $FC_{ij}(\cdot)$ in Eq.~(\ref{eq:pairwise}) is decomposed to two linear layers with a bottleneck dimension $256$ at the middle.
\end{itemize}
This gives supernet better scalability, however the number of choices we can scale is still prohibited when the supernet has full connections among all choices with each choice having all blocks.
We constrain the connections and possible blocks per choice as below and reach a supernet with around one billion dense parameters.

We start from a supernet with an architecture matching our current production model (which is a model variant of DLRM~\cite{naumov2019deep}), where each choice only has one building block to match a layer. We then grow the supernet 
%by adding blocks to each choice module and adding connections among choices, 
as following steps:
\begin{itemize}
\item randomly permute the list of choices while keeping its directed acyclic graph (DAG) order, i.e., predecessors of a node always have smaller indices in the permuted list. This removes the layer order bias encoded by engineers;
\item copy any choice having EmbedFC or compressed dot product block, and then insert it right after;
\item in any choice, if any of [compressed dot product, EmbedFC] block exists, add the other; if any of [linear, pairwise gating, pairwise sum] block exists, add the others;
\item each choice is connected to previous choices with distances of 1, 2, 3, 6 and 9, where ``distance'' is defined as the relative index of two choices in the choice list;
\item enlarge linear dimensions in all blocks to $1.25\times$. There are $7$ dimension options uniformly distributed between $0.5 \times$ and $1.25\times$.
\end{itemize}

The search space consists of $2.8\times 10 ^ {114}$ models, which is $4.3\times 10 ^ {95}$ times larger than the search space size in a typical computer vision problem~\cite{wen2020neural} and $5.6\times 10 ^ {78}$ times larger than a ranking systems NAS work in academia~\cite{zhang2023nasrec}. The number of models is more than the number of atoms in the observable universe~\footnote{https://www.universetoday.com/36302/atoms-in-the-universe/}.

\begin{figure}
\centering
\includegraphics[width=\columnwidth]{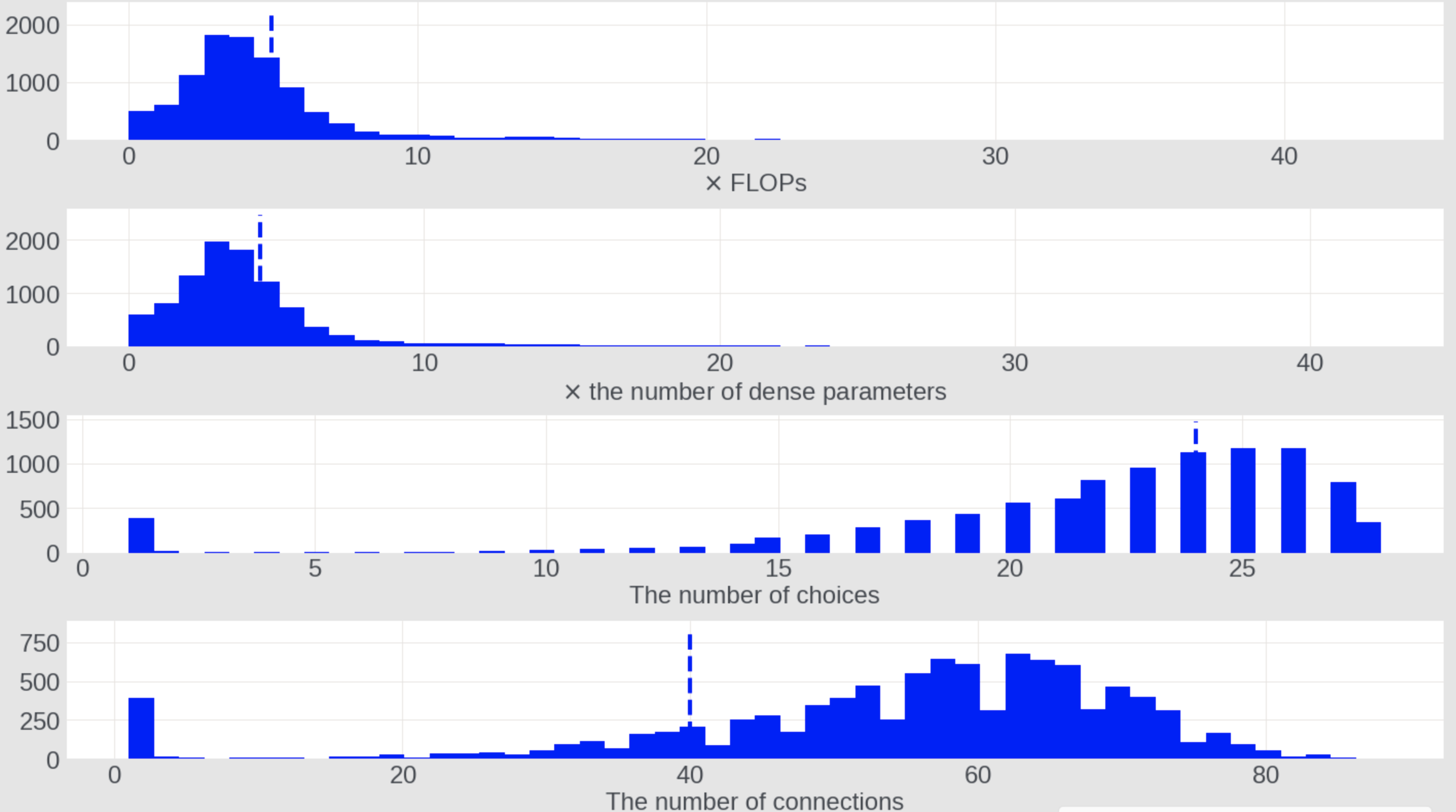}
\caption{The distribution of FLOPs, \# dense parameters, \# choices and \# connections in subnets. Vertical dashed lines are for the production model.}
\label{fig:subnet_hist}
\end{figure}

After finalizing the supernet, we tune the sampling probability of ``K independent binary masks'' in Figure~\ref{fig:supernet} such that the distribution of subnets are reasonable. With probability $0.8$ to set a binary mask to one, Figure~\ref{fig:subnet_hist} plots the histogram of statistics of $10,000$ random subnets, where we can see that distributions spread well around our production model.
Note that the production model is $5 \times$ in terms of FLOPs, which is indicated by vertical dashed lines in Figure~\ref{fig:subnet_hist}.

\subsubsection{Weight-Sharing Proxy and Early-Stop Proxy}\hfill\vspace{0.5em}

\begin{figure}
\centering
\includegraphics[width=\columnwidth]{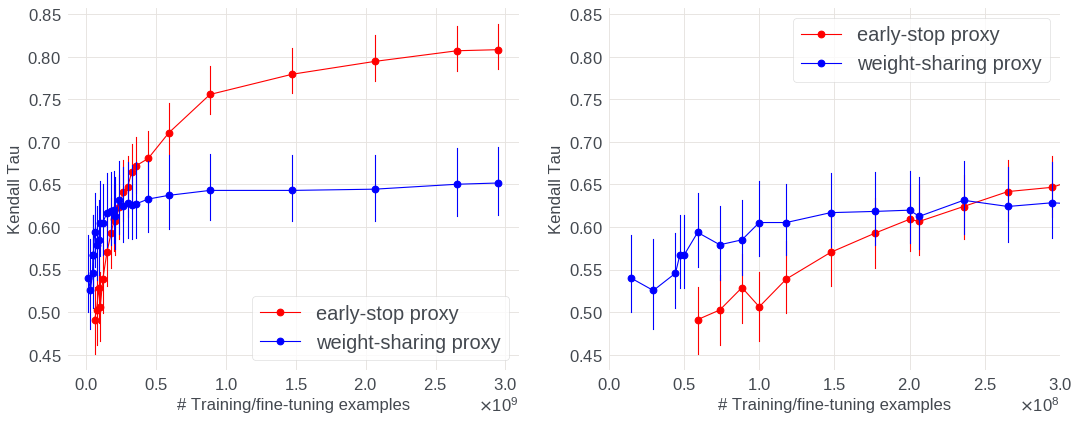}
\caption{Rank correlation between low cost proxies and long term ground truth NE, which is obtained by training for $28$B examples. Vertical intervals are confidence intervals from $25\%$ to $75\%$ quantile. Right figure just zooms into the beginning parts of curves on the left.}
\label{fig:rank_corr}
\end{figure}

To evaluate which low cost NE proxy is better under specific conditions, we randomly sample $60$ models and train each from scratch for $28$B examples (``B'' is for ``billion'') to get long term NE. The window NE averaged on last $0.5$B examples is used as ground truth long term NE.
Figure~\ref{fig:rank_corr} plots the Kendall Tau rank correlation between proxies and ground truth NE.
We find that weight-sharing NE has better correlation quality initially to rank ground truth NE, however, its advantage disappears when the amount of data passes a threshold ($200$ million in our search space).
This is reasonable because weights used in weight-sharing proxy have been pre-trained in the supernet, such that a subnet does not need to train from scratch and therefore outperforms when the training data is less; however, as data amount increases, early-stop proxy approaches to ground truth NE but weight-sharing proxy may have inferior initialization for any subnet to achieve so.

In conclusion, weight-sharing proxy should be used if speedy subnet evaluation is required and a moderate rank quality (e.g. Kendall Tau $0.6$) satisfies; however, early-stop proxy should be used if a high rank correlation is demanded.
Moreover, engineering and machine learning system optimization overheads should also be considered in practice, such as, in weight-sharing proxy, it is non-trivial to implement partial weight transfer from the supernet to any standalone subnet.
In our experiments, we find high rank correlation is important to search models with long term NE, and decided to use early-stop proxy with $3$B training data, which has around $0.8$ Kendall Tau ranking quality.

\subsubsection{Results of Sampling-based Method}\hfill\vspace{0.5em}
\begin{figure}[h]
    % \centering
    \begin{subfigure}[t]{0.5\columnwidth}
        \centering
        \includegraphics[width=\columnwidth]{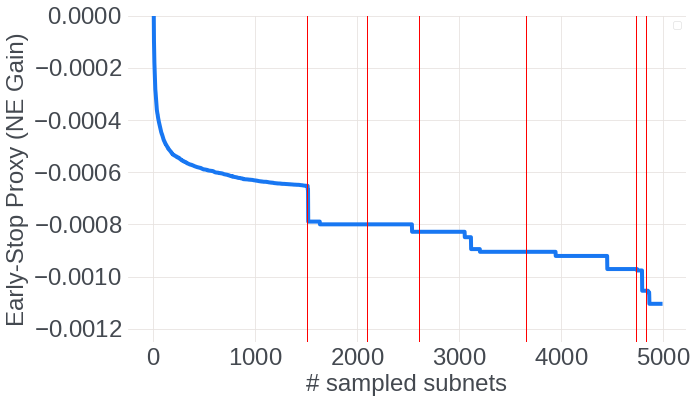}
        \caption{}
        \label{fig:lfe_convergence}
    \end{subfigure}%
    ~
    \begin{subfigure}[t]{0.5\columnwidth}
        \centering
        \includegraphics[width=\columnwidth]{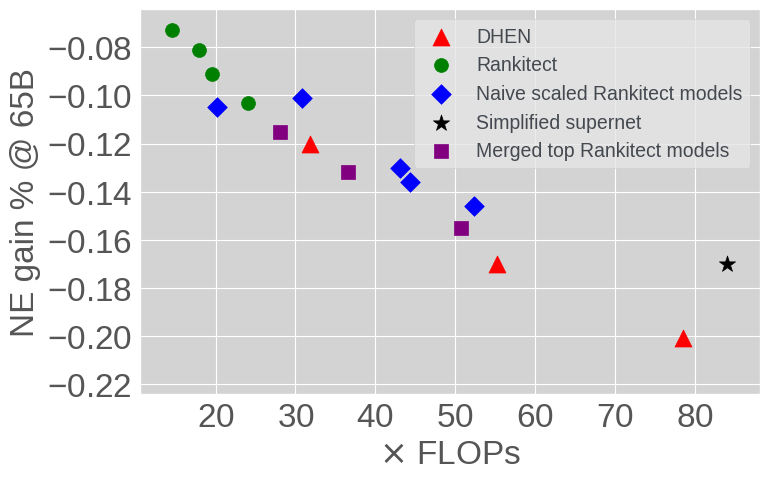}
        \caption{}
        \label{fig:wsnas_vs_5xv0}
    \end{subfigure}
    \caption{Results of Sampling-based Method: (a) NE gain of early-stop proxy. The vertical red lines divide different rounds of sampling-based search. The first round is random search (whose curve is smoothed by averaging over $500$ random permutations), followed by six rounds of Neural Predictor based sampling. For each new round, a MLP is trained by using all previously obtained data points and then picks top models based on MLP prediction; (b) long term NE gain at $65$ billion data examples versus production model.}
\end{figure}

% \begin{figure}
% \centering
% \includegraphics[width=\columnwidth]{Figures/lfe_convergence.png}
% \caption{NE gain of early-stop proxy. The vertical red lines divide different rounds of sampling-based search. The first round is random search (whose curve is smoothed by averaging over $500$ random permutations), followed by six rounds of Neural Predictor based sampling. For each new round, a MLP is trained by using all previously obtained data points and then picks top models based on MLP prediction.}
% \label{fig:lfe_convergence}
% \end{figure}

We applied sampling-based method to search models for low early-stop proxy (which is better if lower). Figure~\ref{fig:lfe_convergence} is the minimal early-stop NE found as more and more models are sampled and evaluated.
In experiments, Neural Predictor bent the curve after initial random sampling. In total, we sampled $5,000$ models. As low early-stop proxy does not guarantee long term NE, we picked $100$ models with lowest early-stop NE and trained for more data with Successive Halving~\cite{jamieson2016non}, ending up with final $4$ best models plotted as green cycles in Figure~\ref{fig:wsnas_vs_5xv0}.

In Figure~\ref{fig:wsnas_vs_5xv0}, we can find that Rankitect can discover a new model with $0.10\%$ absolute NE gain which is considered as significant in ranking systems. Moreover, discovered models cover a good range of model complexity (FLOPs) from $14.5 \times$ to $24 \times$, which provides a good pool of models to select from for different serving cost constraints required by different products and model types.
During Rankitect development, in-house engineers also invented a new model -- DHEN~\cite{zhang2022dhen} as plotted in red triangles. Battling with Rankitect, DHEN covers a new region -- more NE gain with more FLOPs.
The combination of Rankitect and DHEN covers a wider range of FLOPs for different products' serving cost requirements.
To compare with DHEN in the high-FLOPs region, we propose two methods to scale Rankitect models:
\begin{itemize}
    \item Naive scaling (blue diamonds): $3 \times$ linear dimension in EmbedFC (first blue diamond from the left), OR, simultaneously $1.5 \times$ linear dimensions in [linear, EmbedFC and compressed dot product] blocks and $2 \times$ bottleneck dimensions in pairwise gating and sum blocks;
    \item Merging (purple squares): pick a subset of Rankitect models (green circles), for a connection or a block existing in any Rankitect model, add it to the final merged model. For a block having different linear dimensions in different Rankitect models, use the maximal dimension in the final merged model.
\end{itemize}
In Figure~\ref{fig:wsnas_vs_5xv0}, we find the ``Merging'' method has a better NE-FLOPs trade-off than ``Naive scaling''. Battling with DHEN, scaled Rankitect models have on-par NE-FLOPs trade-off trend, but fills the big blank FLOPs region which DHEN is unable to cover.
Moreover, Rankitect automates the process of new architecture design, which can release human engineering resources.

% larger search space to beat it?

% to discovered the following candidates:
% \begin{center}
% \begin{tabular}{c@{\hskip 4em}c@{\hskip 4em}c} 
%  \hline
%  Candidate & NE & FLOPs \\
%  \hline\hline
%  1 & -0.11\% & 20$\times$ \\ 
%  \hline
%  2 & -0.13\% & 37$\times$ \\ 
%  \hline
%  3 & -0.14\% & 44$\times$ \\ 
%  \hline
%  4 & -0.17\% & 84$\times$ \\ 
%  \hline
% \end{tabular}
% \end{center}
% Figure X shows WS-NAS generated candidates compared with state-of-the-art production models created by ML experts.

% \begin{figure}
% \centering
% \includegraphics[width=1.0\columnwidth]{Figures/wsnas_vs_dhen_5x.png}
% \caption{Long term NE gain at $65$ billion data examples versus production model}
% \label{fig:wsnas_vs_5xv0}
% \end{figure}

% From Figure~\ref{fig:wsnas_vs_5xv0}, we can see that WS-NAS can achieve human expert level performance when designing models by machines. This implies that WS-NAS can help to amortize ML HCs by TC/GPU resources.
\begin{table*}[t]
\caption{Rankitect search summary for DHEN-based supernet.}
 \centering
\begin{tabular}{cm{6em}m{9em}cccm{5em}} 
\toprule
 Model & Product applying Rankitect & DHEN layers / \# decisions / search space & Supernet FLOPs & $\alpha$ & Discovered FLOPs & Search cost$^{1}$ (GPU hours) \\
 \hline\hline
%   1 & CTR model-1 & \hspace{2em} 7 / 162 / $9^{162}$ & 140$\times$ & 30$\times$ & 2e-5 & 44.7$\times$ & \hspace{1em} 9216 \\ 
%  \hline
%   2 & CTR model-1 & \hspace{2em} 7 / 162 / $9^{162}$ & 140$\times$ & 30$\times$ & 2e-4 & 26.3$\times$ & \hspace{1em} 9216 \\ 
%  \hline
 1 & CTR app 1 & \hspace{2em} 4 / 93 / $9^{93}$ & 80$\times$  & 2e-5  & 31.2$\times$ & \hspace{1em} 6144 \\ 
 \hline
%  4 & CTR app 1 & \hspace{2em} 4 / 93 / $9^{93}$ & 80$\times$ & 20$\times$ & 2e-4 & 20.3$\times$ & \hspace{1em} 6144 \\ 
%  \hline
 2 & CTR app 1  & \hspace{2em} 2 / 28 / $9^{28}$ & 10$\times$ \iffalse (207M) \fi & 1e-3 & 2.6$\times$ \iffalse (51.8M) \fi & \hspace{1em} 3072 \\ 
 \hline
 3 & CTR app 1 & \hspace{2em} 2 / 28 / $9^{28}$ & 2.65$\times$ \iffalse (53M) \fi  & 2e-3  & 0.6$\times$  \iffalse (12M) \fi & \hspace{1em} 3072 \\ 
 \hline
%  7 & CVR app 2 & \hspace{2em} 2 / 28 / $9^{28}$ & 1.08$\times$ \iffalse (53M) \fi & 0.25$\times$ \iffalse (12M) \fi & 5e-3  & 0.24$\times$ \iffalse (11.9M) \fi & \hspace{1em} 3072 \\ 
%  \hline
 4 & CTR app 1 & \hspace{2em} 1 / 19 / $9^{19}$ & 1.08$\times$  \iffalse (21.7M) \fi & 1e-2  & 0.37$\times$ \iffalse (7.41M) \fi & \hspace{1em} 1536 \\ 
 \bottomrule
%  \multicolumn{8}{l}{\footnotesize 1. relative FLOPs measurement for each candidate are compared against their respective production model type, i.e., not comparable across different production model type.} \\
%  \multicolumn{8}{l}{\footnotesize 2. Max FLOPs indicates the model size with the largest linear operator size greedily selected.} \\
 \multicolumn{7}{l}{\footnotesize 1. all searches ran with 64 GPUs and the search cost are reported in total GPU-hours, e.g., 6144 GPU-hour = 64 GPU * 96 wall-clock hour.}
\end{tabular}
\label{table:dhen_search}
\end{table*}

\subsubsection{Results by One-shot Method and DNAS}\hfill\vspace{0.5em}

\begin{figure}[h]
\centering
\includegraphics[width=0.45\textwidth]{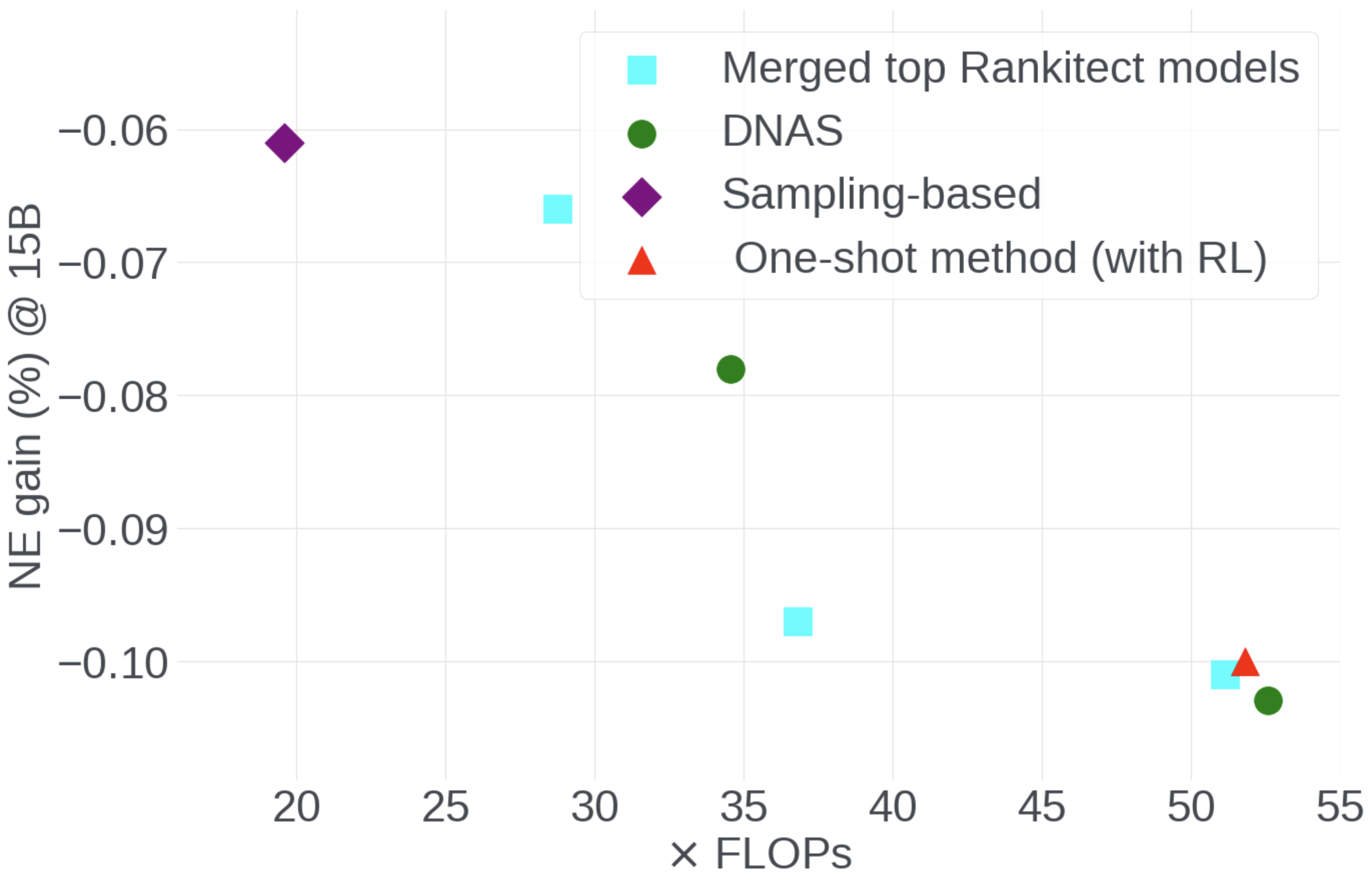}
\caption{NE and FLOPs comparison of three categories of methods.}
\label{fig:dnas_results}
\end{figure}

In our experiments, we find that it is challenging for one-shot method and DNAS to learn connections in the supernet designed in Section~\ref{sec:supernet_design}, therefore, we simplified the supernet by disallowing adding new connections and only learn to select a building block per choice. The NE gain and FLOPs of the simplified supernet is plotted as the black star in Figure~\ref{fig:wsnas_vs_5xv0}. Our goal in this section is to answer the following questions:
\begin{itemize}
    \item Does one-shot method or DNAS produce better NE models than sampling-based method?
    \item How do the search costs of three categories of methods compare?
    \item Is DNAS able to leverage the reduced variance of the Gumbel-Softmax gradient estimator compared to RL's policy gradient (in one-shot method) to converge to solutions which satisfy a FLOPs constraint faster?
\end{itemize}
Figure~\ref{fig:dnas_results} shows that the DNAS is able to find models which outperform the production model and sampling-based model, albeit at higher FLOPs cost. When we looked at the efficiency/cost of each method, DNAS and RL yield results with roughly $145 \times$ less compute resources ($1.45$ days vs. $209.5$ days used by sampling-based method on 64 GPU system), which is a huge benefit at Meta-scale recommendation system training. Finally, we compare DNAS to RL and evaluate the effect of gradient variance on convergence performance since one of the benefits of DNAS is the reduced gradient variance compared to RL \cite{jang2016categorical}. We setup an experiment where both DNAS and RL try to search for a model with a target FLOPs (without considering its NE) and report that DNAS requires $\sim 26\times$ less SGD steps to converge to a given FLOPs target.

% \begin{figure}[h]
% \centering
% \includegraphics[width=0.47\textwidth]{Figures/dnas_search_cost.png}
% \caption{Search cost of DNAS, RL, and predictor-based systems.}
% \label{fig:dnas_search_cost}
% \end{figure}

% \begin{figure}[h]
% \centering
% \includegraphics[width=0.47\textwidth]{Figures/dnas_flops_constraint.png}
% \caption{Number of steps required by DNAS and REINFORCE-based algorithms to converge to solution within $1\%$ of the FLOPs target}
% \label{fig:dnas_flops_constraint}
% \end{figure}

Although DNAS has overall the best performance, in the end we chose to focus our software system on RL instead of DNAS for the following reasons:
\begin{itemize}
    \item The engineering cost of maintaining two search algorithms with relatively similar search performance is too high.
    \item Memory consumption of naive DNAS implementation is higher, demanding potentially more memory optimization overhead.
    \item The engineering cost of DNAS is higher than RL. DNAS requires careful tuning of hyperparameters like Gumbel-Softmax temperature, sparsity level of sampled r.v's, etc. \cite{NEURIPS2022_753d9584}. Since the loss function is required to be differentiable for DNAS, optimization of FLOPs requires us to build a software library modeling FLOPs as a differentiable function of architecture. While this is possible, the engineering cost of building and maintaining such a library are quite high. On the other hand, RL can natively optimize non-differentiable objectives.  
\end{itemize}

\subsection{Reusing Search Space Invented by World-class Engineers}

% \begin{table}%[!htb]
%     \caption{Generalization of Rankitect discovered models.}
%     \begin{minipage}{1\linewidth}
% \centering
% \begin{tabular}{c@{\hskip 4em}c@{\hskip 4em}c} 
% \multicolumn{3}{c}{\textbf{Model 6 performance across products}} \\[5pt]
%  \toprule
%  Products & NE gain & FLOPs \\
%  \hline\hline
%  CVR app 1 & -0.053\% & -34.5\% \\ 
%  \hline
%  CVR app 2 & 0.04\% & -40\% \\ 
%  \hline
%  CVR app 3 & -0.099\% & -31.5\% \\ 
%  \bottomrule
%  \label{table:smt}
% \end{tabular}
%     \end{minipage}
%     \vspace{1em}
%     \begin{minipage}{1\linewidth}
%       \centering
% \begin{tabular}{c@{\hskip 4em}c@{\hskip 4em}c} 
% \multicolumn{3}{c}{\textbf{Model 8 performance across products}} \\[5pt]
%  \toprule
%  Products & NE gain & FLOPs \\
%  \hline\hline
%  CVR app 1 & -0.173\% & +20\% \\ 
%  \hline
%  CVR app 2 & -0.25\% & +7\% \\ 
%  \hline
%  CVR app 3 & -0.254\% & +23\% \\ 
%  \bottomrule
% \end{tabular}
%     \end{minipage} 
% \end{table}

As proved in Section~\ref{sec:full_space_exp}, Rankitect automatically produces competitive models versus world-class engineers with more diverse FLOPs coverage. More importantly, the result is generated from an unspecified supernet (or search space) with minor human prior.
To bring the advantages from both worlds (of Rankitect and in-house engineers) together, we use DHEN as supernet backbone and apply Rankitect.
We hypothesize that: human crafted DHEN-based supernet has a higher density of good models than the unspecified supernet, which makes Rankitect easier to find better models and can outperform human experts. We prove so in this section.

\iffalse In the previous section we applied our Rankitect framework to a large search space and although we show that Rankitect can generate strong candidate models with significant NE gain, searching over a large search space is both technically and computationally challenging and reduces its ROI. To realize Rankitect in production and increase its ROI
\fi 
% To further increase the ROI of applying Rankitect NAS in production, we reuse a search space inspired by in-house expert engineers. 
We setup a supernet to match the architecture of Deep and Hierarchical Ensemble Networks (DHEN) \cite{zhang2022dhen} and focus on searching for all building block dimensions. For example, a $4$ layer DHEN-based supernet has $93$ building block dimensions to search; with each block dimension having 9 options,  we have a search space with size $9^{93}$. The primary motivation for us to only search for block dimensions is to ensure an easy transfer of Rankitect search results to DHEN models in Meta's production stack. 
% For example, we can identify the relevant production model settings (such as various MLP bottleneck projection) and directly transfer Rankitect size search results. 
To test the transferability, Rankitect is only applied on the strongest baseline at Meta for the ``CTR app 1'' product, and we simply apply the discovered models to other products.
To meet different serving cost constraints of different products, we applied our Rankitect framework to various DHEN-based supernets, targeting different FLOPs goals.
We summarize our search results in Table~\ref{table:dhen_search}.
% where candidates 3, 4, 5, 6, 8 satisfied various production needs at the time and were evaluated further. 
% We summarize their production results in the following: 
% \iffalse Note that in the RL search, we can amortize part of the search cost by reusing supernet warm-up stage as long as they share the same search space and model type. For example, in Table~\ref{table:dhen_search} the warm-up stage of 7-layer (4-layer, 2-layer) DHEN-based supernet takes around 3 (2, 1) days but can be reused for all subsequent search with the same supernet, e.g., candidate 1 and 2 (3 and 4, 5 and 6) can share the same supernet warm-up stage (for this reason we did not include the supernet warm-up cost in the reported search cost). From the search results in Table~\ref{table:dhen_search}, candidates 3, 4, 5, 6, 8 satisfied various production needs at the time and were evaluated further. We summarize their performance in the following: \fi

% \vspace{0.5em}
\paragraph{\textbf{Beating strongest human baseline}}
During model iteration of ``CTR app 1'', our in-house engineers were targeting on an approximately $30 \times$ model for product and established the largest and strongest CTR baseline at Meta. To battle human engineers, Rankitect generated ``model-1'' in Table~\ref{table:dhen_search} which is a $31.2\times$ model with $> 0.02\%$ absolute NE gain at $90$B (with gain still enlarging with more training data shown in Figure~\ref{fig:wsnas_vs_dhen31x} in Appendix) compared with the human tuned $31.3\times$ baseline model. %Figure~\ref{fig:wsnas_vs_dhen31x} in Appendix shows the offline training NE-gain of candidate 3 compared against a 31.3$\times$ model tuned by strong human (both model train-from-scratch).
This proves that Rankitect can discover better models than Meta's world-class engineers. 

% \vspace{0.5em}
% \noindent\textbf{\textbullet\ Candidate 4 is a 20.3$\times$ model with 0.06\% NE-gain @50B (and trending better) compared with production 10$\times$ baseline model.} 
% % Figure~\ref{fig:wsnas_vs_10xV0} shows the offline training NE-gain (compared against production 10$\times$ model) of candidate 4 when train-from-scratch. 
% In contrast, a 18.5$\times$ model designed by strong human has a flat 0.05\% NE-gain when compared against the same baseline. Again, this demonstrated that Rankitect can reduce engineers by machines at production scale

% \begin{figure}[h]
% \centering
% \includegraphics[width=0.48\textwidth]{Figures/wsnas_vs_10xV0.png}
% \caption{Offline training NE-gain of candidate 4 compared against production 10$\times$ model.}
% \label{fig:wsnas_vs_10xV0}
% \end{figure}
\begin{table}
\caption{Comparison of Rankitect and sampling-based production AutoML searcher.}
\centering
\begin{tabular}{m{5em}c@{\hskip 1em}c@{\hskip 1em}m{5em}}
 \toprule
 Method & Model size & NE gain & Search cost (GPU-hour) \\
 \hline\hline
 Rankitect & 2.6$\times$ \iffalse (51.8M) \fi & -0.28\% & \hspace{1em} 3072 \\ 
 \hline
 Production AutoML & 3.14$\times$ \iffalse (62.6M) \fi & -0.35\% &  \hspace{1em} 25600 \\ 
 \bottomrule
\end{tabular}
\label{table:rankitect_vs_sampling}
\end{table}

\paragraph{\textbf{Beating AutoML in production}}
We applied ``model 2'' to product ``CVR app 0'' and conducted a side-by-side comparison with a model discovered by sampling-based AutoML in production shown in Table~\ref{table:rankitect_vs_sampling}.
It turns out that our Rankitect model was able to meet FLOPs constraint for this product with significant NE gain, while production AutoML was not able to satisfy the FLOPs constraint.
Furthermore, Rankitect achieved $8.3\times$ search efficiency gain over production AutoML (measured by GPU hours). Due to those, ``model 2'' was selected for online A/B test and show \emph{statistically significant} gain over production model. 
%In summary, Rankitect are both more effective (discover smaller model with similar NE) and efficient (8.3$\times$ more search efficient) than a strong in-house sampling-based method.

% It shows that candidate 5---a 2.6$\times$ model---achieved similar NE performance compared to a 3.14$\times$ model discovered by a competing in-house sampling-based method. Moreover, Rankitect spent 3072 GPU-hours for the search compared to 25600 GPU-hour used by the sampling-based method leading to a 8.3$\times$ search efficiency gain
% Furthermore, the search cost of discovering candidate 5 by Rankitect is 8.3$\times$ more TC efficient than the sampling-based searcher. A detailed comparison is summarized in Table~\ref{table:rankitect_vs_sampling}.
% \begin{table}[h]
% \caption{Search comparison of Rankitect and sampling-based production AutoML searcher.}
% \centering
% \begin{tabular}{m{10em}m{6em}m{5.5em}}
%  \hline
%  Search method & Discovered model size & Search cost (GPU-hour) \\
%  \hline\hline
%  Rankitect & 1.06$\times$ \iffalse (51.8M) \fi & 3072 \\ 
%  \hline
%  Sampling based searcher & 1.28$\times$ \iffalse (62.6M) \fi & 25600 \\ 
%  \hline
% \end{tabular}
% \label{table:rankitect_vs_sampling}
% \end{table}

\vspace{0.5em}
% \noindent\textbf{\textbullet\ Candidate 6 and 8 demonstrated that Rankitect can search for once competitive model once and applied the discovered model to many applications (search-once-apply-many)}. When applying Rankitect to search for different model types, the ideal case is to conduct the search using the target model type’s input data pipeline; however, this will require us to migrate/maintain a new data pipeline to Rankitect every time we search a new model type. In this work, we adopted the strategy of searching using the same input data pipeline for all model types but compressed the input (dense and sparse) data shape to match that of the target model type. This minimizes the engineering overhead while still allowing us to apply Rankitect to search for new model type. This strategy were used for all the candidate search in Table~\ref{table:dhen_search}. Next, we applied the discovered candidate 6 and 8 across different production model types (including model types different from the original search target model type) and demonstrated their competitive performance. The performance of candidate 6 and 8 compared against production baseline over various models are shown in Table~\ref{table:cand6_smt} and \ref{table:cand8_smt}, respectively.
\paragraph{\textbf{Strong model transferability to other products}}
We applied ``Model 3'' and ``Model 4'' across different products and observe strong model transferability. The results are summarized in Table~\ref{table:smt} where we can see that ``Model 3'' is a smaller model but still achieve better NE performance in many of the products. On the other hand, ``Model 4'' achieve significant NE-gain over product baselines at the cost of model size increase. These results demonstrate that Rankitect has the potential to search once and apply to many.
\begin{table}
    \caption{Transferability of Rankitect discovered models.}
    \begin{tabular}{c@{\hskip 2em}c@{\hskip 3em}c@{\hskip 3em}c} 
 \toprule
 Model & Products & NE gain & FLOPs \\
 \hline\hline
 3 & CVR app 1 & -0.053\% & -34.5\% \\ 
 \hline
 3 & CVR app 2 & 0.04\% & -40\% \\ 
 \hline
 3 & CVR app 3 & -0.099\% & -31.5\% \\ 
 \hline\hline
 4 & CVR app 1 & -0.173\% & +20\% \\ 
 \hline
 4 & CVR app 2 & -0.25\% & +7\% \\ 
 \hline
 4 & CVR app 3 & -0.254\% & +23\% \\ 
 \bottomrule
 \label{table:smt}
\end{tabular}
\end{table}
\newpage

%Bibliography
\bibliographystyle{ACM-Reference-Format}  
\bibliography{main}  


\begin{thebibliography}{46}

%%% the Web address.

\ifx \showCODEN    \undefined \def \showCODEN     #1{\unskip}     \fi
\ifx \showDOI      \undefined \def \showDOI       #1{#1}\fi
\ifx \showISBNx    \undefined \def \showISBNx     #1{\unskip}     \fi
\ifx \showISBNxiii \undefined \def \showISBNxiii  #1{\unskip}     \fi
\ifx \showISSN     \undefined \def \showISSN      #1{\unskip}     \fi
\ifx \showLCCN     \undefined \def \showLCCN      #1{\unskip}     \fi
\ifx \shownote     \undefined \def \shownote      #1{#1}          \fi
\ifx \showarticletitle \undefined \def \showarticletitle #1{#1}   \fi
\ifx \showURL      \undefined \def \showURL       {\relax}        \fi
% invisible to TeX
\providecommand\bibfield[2]{#2}
\providecommand\bibinfo[2]{#2}
\providecommand\natexlab[1]{#1}
\providecommand\showeprint[2][]{arXiv:#2}

\bibitem[Abdelfattah et~al\mbox{.}(2021)]%
        {abdelfattah2021zero}
\bibfield{author}{\bibinfo{person}{Mohamed~S Abdelfattah},
  \bibinfo{person}{Abhinav Mehrotra}, \bibinfo{person}{{\L}ukasz Dudziak},
  {and} \bibinfo{person}{Nicholas~D Lane}.} \bibinfo{year}{2021}\natexlab{}.
\newblock \showarticletitle{Zero-cost proxies for lightweight nas}.
\newblock \bibinfo{journal}{\emph{arXiv preprint arXiv:2101.08134}}
  (\bibinfo{year}{2021}).
\newblock


\bibitem[Anil et~al\mbox{.}(2022)]%
        {anil2022factory}
\bibfield{author}{\bibinfo{person}{Rohan Anil}, \bibinfo{person}{Sandra
  Gadanho}, \bibinfo{person}{Da Huang}, \bibinfo{person}{Nijith Jacob},
  \bibinfo{person}{Zhuoshu Li}, \bibinfo{person}{Dong Lin},
  \bibinfo{person}{Todd Phillips}, \bibinfo{person}{Cristina Pop},
  \bibinfo{person}{Kevin Regan}, \bibinfo{person}{Gil~I Shamir},
  {et~al\mbox{.}}} \bibinfo{year}{2022}\natexlab{}.
\newblock \showarticletitle{On the factory floor: ML engineering for
  industrial-scale ads recommendation models}.
\newblock \bibinfo{journal}{\emph{arXiv preprint arXiv:2209.05310}}
  (\bibinfo{year}{2022}).
\newblock


\bibitem[Banbury et~al\mbox{.}(2021)]%
        {MLSYS2021_c4d41d96}
\bibfield{author}{\bibinfo{person}{Colby Banbury}, \bibinfo{person}{Chuteng
  Zhou}, \bibinfo{person}{Igor Fedorov}, \bibinfo{person}{Ramon Matas},
  \bibinfo{person}{Urmish Thakker}, \bibinfo{person}{Dibakar Gope},
  \bibinfo{person}{Vijay Janapa~Reddi}, \bibinfo{person}{Matthew Mattina},
  {and} \bibinfo{person}{Paul Whatmough}.} \bibinfo{year}{2021}\natexlab{}.
\newblock \showarticletitle{MicroNets: Neural Network Architectures for
  Deploying TinyML Applications on Commodity Microcontrollers}. In
  \bibinfo{booktitle}{\emph{Proceedings of Machine Learning and Systems}},
  \bibfield{editor}{\bibinfo{person}{A.~Smola}, \bibinfo{person}{A.~Dimakis},
  {and} \bibinfo{person}{I.~Stoica}} (Eds.), Vol.~\bibinfo{volume}{3}.
  \bibinfo{pages}{517--532}.
\newblock
\urldef\tempurl%
\url{https://proceedings.mlsys.org/paper_files/paper/2021/file/c4d41d9619462c534b7b61d1f772385e-Paper.pdf}
\showURL{%
\tempurl}


\bibitem[Bender et~al\mbox{.}(2018)]%
        {bender2018understanding}
\bibfield{author}{\bibinfo{person}{Gabriel Bender}, \bibinfo{person}{Pieter-Jan
  Kindermans}, \bibinfo{person}{Barret Zoph}, \bibinfo{person}{Vijay
  Vasudevan}, {and} \bibinfo{person}{Quoc Le}.}
  \bibinfo{year}{2018}\natexlab{}.
\newblock \showarticletitle{Understanding and simplifying one-shot architecture
  search}. In \bibinfo{booktitle}{\emph{International conference on machine
  learning}}. PMLR, \bibinfo{pages}{550--559}.
\newblock


\bibitem[Bender et~al\mbox{.}(2020a)]%
        {9157751}
\bibfield{author}{\bibinfo{person}{G. Bender}, \bibinfo{person}{H. Liu},
  \bibinfo{person}{B. Chen}, \bibinfo{person}{G. Chu}, \bibinfo{person}{S.
  Cheng}, \bibinfo{person}{P. Kindermans}, {and} \bibinfo{person}{Q.~V. Le}.}
  \bibinfo{year}{2020}\natexlab{a}.
\newblock \showarticletitle{Can Weight Sharing Outperform Random Architecture
  Search? An Investigation With TuNAS}. In \bibinfo{booktitle}{\emph{2020
  IEEE/CVF Conference on Computer Vision and Pattern Recognition (CVPR)}}.
  \bibinfo{publisher}{IEEE Computer Society}, \bibinfo{address}{Los Alamitos,
  CA, USA}, \bibinfo{pages}{14311--14320}.
\newblock
\urldef\tempurl%
\url{https://doi.org/10.1109/CVPR42600.2020.01433}
\showDOI{\tempurl}


\bibitem[Bender et~al\mbox{.}(2020b)]%
        {bender2020can}
\bibfield{author}{\bibinfo{person}{Gabriel Bender}, \bibinfo{person}{Hanxiao
  Liu}, \bibinfo{person}{Bo Chen}, \bibinfo{person}{Grace Chu},
  \bibinfo{person}{Shuyang Cheng}, \bibinfo{person}{Pieter-Jan Kindermans},
  {and} \bibinfo{person}{Quoc~V Le}.} \bibinfo{year}{2020}\natexlab{b}.
\newblock \showarticletitle{Can weight sharing outperform random architecture
  search? an investigation with tunas}. In
  \bibinfo{booktitle}{\emph{Proceedings of the IEEE/CVF Conference on Computer
  Vision and Pattern Recognition}}. \bibinfo{pages}{14323--14332}.
\newblock


\bibitem[Cai et~al\mbox{.}(2018)]%
        {cai2018proxylessnas}
\bibfield{author}{\bibinfo{person}{Han Cai}, \bibinfo{person}{Ligeng Zhu},
  {and} \bibinfo{person}{Song Han}.} \bibinfo{year}{2018}\natexlab{}.
\newblock \showarticletitle{ProxylessNAS: Direct Neural Architecture Search on
  Target Task and Hardware}. In \bibinfo{booktitle}{\emph{International
  Conference on Learning Representations}}.
\newblock


\bibitem[Cao et~al\mbox{.}(2023)]%
        {cao2023farthest}
\bibfield{author}{\bibinfo{person}{Yufan Cao}, \bibinfo{person}{Tunhou Zhang},
  \bibinfo{person}{Wei Wen}, \bibinfo{person}{Feng Yan}, \bibinfo{person}{Hai
  Li}, {and} \bibinfo{person}{Yiran Chen}.} \bibinfo{year}{2023}\natexlab{}.
\newblock \showarticletitle{Farthest Greedy Path Sampling for Two-shot
  Recommender Search}.
\newblock \bibinfo{journal}{\emph{arXiv preprint arXiv:2310.20705}}
  (\bibinfo{year}{2023}).
\newblock


\bibitem[Dong and Yang(2019)]%
        {Dong_2019_CVPR}
\bibfield{author}{\bibinfo{person}{Xuanyi Dong} {and} \bibinfo{person}{Yi
  Yang}.} \bibinfo{year}{2019}\natexlab{}.
\newblock \showarticletitle{Searching for a Robust Neural Architecture in Four
  GPU Hours}. In \bibinfo{booktitle}{\emph{Proceedings of the IEEE/CVF
  Conference on Computer Vision and Pattern Recognition (CVPR)}}.
\newblock


\bibitem[Fedorov et~al\mbox{.}(2019)]%
        {NEURIPS2019_044a23ca}
\bibfield{author}{\bibinfo{person}{Igor Fedorov}, \bibinfo{person}{Ryan~P
  Adams}, \bibinfo{person}{Matthew Mattina}, {and} \bibinfo{person}{Paul
  Whatmough}.} \bibinfo{year}{2019}\natexlab{}.
\newblock \showarticletitle{SpArSe: Sparse Architecture Search for CNNs on
  Resource-Constrained Microcontrollers}. In \bibinfo{booktitle}{\emph{Advances
  in Neural Information Processing Systems}},
  \bibfield{editor}{\bibinfo{person}{H.~Wallach},
  \bibinfo{person}{H.~Larochelle}, \bibinfo{person}{A.~Beygelzimer},
  \bibinfo{person}{F.~d\textquotesingle Alch\'{e}-Buc},
  \bibinfo{person}{E.~Fox}, {and} \bibinfo{person}{R.~Garnett}} (Eds.),
  Vol.~\bibinfo{volume}{32}. \bibinfo{publisher}{Curran Associates, Inc.}
\newblock
\urldef\tempurl%
\url{https://proceedings.neurips.cc/paper_files/paper/2019/file/044a23cadb567653eb51d4eb40acaa88-Paper.pdf}
\showURL{%
\tempurl}


\bibitem[Fedorov et~al\mbox{.}(2022)]%
        {NEURIPS2022_753d9584}
\bibfield{author}{\bibinfo{person}{Igor Fedorov}, \bibinfo{person}{Ramon
  Matas}, \bibinfo{person}{Hokchhay Tann}, \bibinfo{person}{Chuteng Zhou},
  \bibinfo{person}{Matthew Mattina}, {and} \bibinfo{person}{Paul Whatmough}.}
  \bibinfo{year}{2022}\natexlab{}.
\newblock \showarticletitle{UDC: Unified DNAS for Compressible TinyML Models
  for Neural Processing Units}. In \bibinfo{booktitle}{\emph{Advances in Neural
  Information Processing Systems}},
  \bibfield{editor}{\bibinfo{person}{S.~Koyejo}, \bibinfo{person}{S.~Mohamed},
  \bibinfo{person}{A.~Agarwal}, \bibinfo{person}{D.~Belgrave},
  \bibinfo{person}{K.~Cho}, {and} \bibinfo{person}{A.~Oh}} (Eds.),
  Vol.~\bibinfo{volume}{35}. \bibinfo{publisher}{Curran Associates, Inc.},
  \bibinfo{pages}{18456--18471}.
\newblock
\urldef\tempurl%
\url{https://proceedings.neurips.cc/paper_files/paper/2022/file/753d9584b57ba01a10482f1ea7734a89-Paper-Conference.pdf}
\showURL{%
\tempurl}


\bibitem[Gao et~al\mbox{.}(2021)]%
        {gao2021progressive}
\bibfield{author}{\bibinfo{person}{Chen Gao}, \bibinfo{person}{Yinfeng Li},
  \bibinfo{person}{Quanming Yao}, \bibinfo{person}{Depeng Jin}, {and}
  \bibinfo{person}{Yong Li}.} \bibinfo{year}{2021}\natexlab{}.
\newblock \showarticletitle{Progressive feature interaction search for deep
  sparse network}.
\newblock \bibinfo{journal}{\emph{Advances in Neural Information Processing
  Systems}}  \bibinfo{volume}{34} (\bibinfo{year}{2021}),
  \bibinfo{pages}{392--403}.
\newblock


\bibitem[He et~al\mbox{.}(2014)]%
        {he2014practical}
\bibfield{author}{\bibinfo{person}{Xinran He}, \bibinfo{person}{Junfeng Pan},
  \bibinfo{person}{Ou Jin}, \bibinfo{person}{Tianbing Xu}, \bibinfo{person}{Bo
  Liu}, \bibinfo{person}{Tao Xu}, \bibinfo{person}{Yanxin Shi},
  \bibinfo{person}{Antoine Atallah}, \bibinfo{person}{Ralf Herbrich},
  \bibinfo{person}{Stuart Bowers}, {et~al\mbox{.}}}
  \bibinfo{year}{2014}\natexlab{}.
\newblock \showarticletitle{Practical lessons from predicting clicks on ads at
  facebook}. In \bibinfo{booktitle}{\emph{Proceedings of the eighth
  international workshop on data mining for online advertising}}.
  \bibinfo{pages}{1--9}.
\newblock


\bibitem[Jamieson and Talwalkar(2016)]%
        {jamieson2016non}
\bibfield{author}{\bibinfo{person}{Kevin Jamieson} {and} \bibinfo{person}{Ameet
  Talwalkar}.} \bibinfo{year}{2016}\natexlab{}.
\newblock \showarticletitle{Non-stochastic best arm identification and
  hyperparameter optimization}. In \bibinfo{booktitle}{\emph{Artificial
  intelligence and statistics}}. PMLR, \bibinfo{pages}{240--248}.
\newblock


\bibitem[Jang et~al\mbox{.}(2016)]%
        {jang2016categorical}
\bibfield{author}{\bibinfo{person}{Eric Jang}, \bibinfo{person}{Shixiang Gu},
  {and} \bibinfo{person}{Ben Poole}.} \bibinfo{year}{2016}\natexlab{}.
\newblock \showarticletitle{Categorical Reparameterization with
  Gumbel-Softmax}. In \bibinfo{booktitle}{\emph{International Conference on
  Learning Representations}}.
\newblock


\bibitem[Kingma and Welling(2022)]%
        {kingma2022autoencoding}
\bibfield{author}{\bibinfo{person}{Diederik~P Kingma} {and}
  \bibinfo{person}{Max Welling}.} \bibinfo{year}{2022}\natexlab{}.
\newblock \bibinfo{title}{Auto-Encoding Variational Bayes}.
\newblock
\newblock
\showeprint[arxiv]{1312.6114}~[stat.ML]


\bibitem[Klyuchnikov et~al\mbox{.}(2022)]%
        {klyuchnikov2022bench}
\bibfield{author}{\bibinfo{person}{Nikita Klyuchnikov}, \bibinfo{person}{Ilya
  Trofimov}, \bibinfo{person}{Ekaterina Artemova}, \bibinfo{person}{Mikhail
  Salnikov}, \bibinfo{person}{Maxim Fedorov}, \bibinfo{person}{Alexander
  Filippov}, {and} \bibinfo{person}{Evgeny Burnaev}.}
  \bibinfo{year}{2022}\natexlab{}.
\newblock \showarticletitle{Nas-bench-nlp: neural architecture search benchmark
  for natural language processing}.
\newblock \bibinfo{journal}{\emph{IEEE Access}}  \bibinfo{volume}{10}
  (\bibinfo{year}{2022}), \bibinfo{pages}{45736--45747}.
\newblock


\bibitem[Krishna et~al\mbox{.}(2021)]%
        {krishna2021differentiable}
\bibfield{author}{\bibinfo{person}{Ravi Krishna}, \bibinfo{person}{Aravind
  Kalaiah}, \bibinfo{person}{Bichen Wu}, \bibinfo{person}{Maxim Naumov},
  \bibinfo{person}{Dheevatsa Mudigere}, \bibinfo{person}{Misha Smelyanskiy},
  {and} \bibinfo{person}{Kurt Keutzer}.} \bibinfo{year}{2021}\natexlab{}.
\newblock \showarticletitle{Differentiable NAS Framework and Application to Ads
  CTR Prediction}.
\newblock \bibinfo{journal}{\emph{arXiv preprint arXiv:2110.14812}}
  (\bibinfo{year}{2021}).
\newblock


\bibitem[Li et~al\mbox{.}(2023)]%
        {li2023hyperscale}
\bibfield{author}{\bibinfo{person}{Sheng Li}, \bibinfo{person}{Garrett
  Andersen}, \bibinfo{person}{Tao Chen}, \bibinfo{person}{Liqun Cheng},
  \bibinfo{person}{Julian Grady}, \bibinfo{person}{Da Huang},
  \bibinfo{person}{Quoc~V Le}, \bibinfo{person}{Andrew Li},
  \bibinfo{person}{Xin Li}, \bibinfo{person}{Yang Li}, {et~al\mbox{.}}}
  \bibinfo{year}{2023}\natexlab{}.
\newblock \showarticletitle{Hyperscale Hardware Optimized Neural Architecture
  Search}. In \bibinfo{booktitle}{\emph{Proceedings of the 28th ACM
  International Conference on Architectural Support for Programming Languages
  and Operating Systems, Volume 3}}. \bibinfo{pages}{343--358}.
\newblock


\bibitem[Li et~al\mbox{.}(2020)]%
        {li2020pytorch}
\bibfield{author}{\bibinfo{person}{Shen Li}, \bibinfo{person}{Yanli Zhao},
  \bibinfo{person}{Rohan Varma}, \bibinfo{person}{Omkar Salpekar},
  \bibinfo{person}{Pieter Noordhuis}, \bibinfo{person}{Teng Li},
  \bibinfo{person}{Adam Paszke}, \bibinfo{person}{Jeff Smith},
  \bibinfo{person}{Brian Vaughan}, \bibinfo{person}{Pritam Damania}, {and}
  \bibinfo{person}{Soumith Chintala}.} \bibinfo{year}{2020}\natexlab{}.
\newblock \bibinfo{title}{PyTorch Distributed: Experiences on Accelerating Data
  Parallel Training}.
\newblock
\newblock
\showeprint[arxiv]{2006.15704}~[cs.DC]


\bibitem[Liu et~al\mbox{.}(2020)]%
        {liu2020autofis}
\bibfield{author}{\bibinfo{person}{Bin Liu}, \bibinfo{person}{Chenxu Zhu},
  \bibinfo{person}{Guilin Li}, \bibinfo{person}{Weinan Zhang},
  \bibinfo{person}{Jincai Lai}, \bibinfo{person}{Ruiming Tang},
  \bibinfo{person}{Xiuqiang He}, \bibinfo{person}{Zhenguo Li}, {and}
  \bibinfo{person}{Yong Yu}.} \bibinfo{year}{2020}\natexlab{}.
\newblock \showarticletitle{Autofis: Automatic feature interaction selection in
  factorization models for click-through rate prediction}. In
  \bibinfo{booktitle}{\emph{proceedings of the 26th ACM SIGKDD international
  conference on knowledge discovery \& data mining}}.
  \bibinfo{pages}{2636--2645}.
\newblock


\bibitem[Liu et~al\mbox{.}(2018)]%
        {liu2018darts}
\bibfield{author}{\bibinfo{person}{Hanxiao Liu}, \bibinfo{person}{Karen
  Simonyan}, {and} \bibinfo{person}{Yiming Yang}.}
  \bibinfo{year}{2018}\natexlab{}.
\newblock \showarticletitle{Darts: Differentiable architecture search}.
\newblock \bibinfo{journal}{\emph{arXiv preprint arXiv:1806.09055}}
  (\bibinfo{year}{2018}).
\newblock


\bibitem[Mahmood and Sutton(2015)]%
        {10.5555/3020847.3020905}
\bibfield{author}{\bibinfo{person}{A.~Rupam Mahmood} {and}
  \bibinfo{person}{Richard~S. Sutton}.} \bibinfo{year}{2015}\natexlab{}.
\newblock \showarticletitle{Off-Policy Learning Based on Weighted Importance
  Sampling with Linear Computational Complexity}. In
  \bibinfo{booktitle}{\emph{Proceedings of the Thirty-First Conference on
  Uncertainty in Artificial Intelligence}} (Amsterdam, Netherlands)
  \emph{(\bibinfo{series}{UAI'15})}. \bibinfo{publisher}{AUAI Press},
  \bibinfo{address}{Arlington, Virginia, USA}, \bibinfo{pages}{552–561}.
\newblock
\showISBNx{9780996643108}


\bibitem[Mehrotra et~al\mbox{.}(2020)]%
        {mehrotra2020bench}
\bibfield{author}{\bibinfo{person}{Abhinav Mehrotra}, \bibinfo{person}{Alberto
  Gil~CP Ramos}, \bibinfo{person}{Sourav Bhattacharya},
  \bibinfo{person}{{\L}ukasz Dudziak}, \bibinfo{person}{Ravichander Vipperla},
  \bibinfo{person}{Thomas Chau}, \bibinfo{person}{Mohamed~S Abdelfattah},
  \bibinfo{person}{Samin Ishtiaq}, {and} \bibinfo{person}{Nicholas~Donald
  Lane}.} \bibinfo{year}{2020}\natexlab{}.
\newblock \showarticletitle{NAS-Bench-ASR: Reproducible neural architecture
  search for speech recognition}. In \bibinfo{booktitle}{\emph{International
  Conference on Learning Representations}}.
\newblock


\bibitem[Metelli et~al\mbox{.}(2020)]%
        {JMLR:v21:20-124}
\bibfield{author}{\bibinfo{person}{Alberto~Maria Metelli},
  \bibinfo{person}{Matteo Papini}, \bibinfo{person}{Nico Montali}, {and}
  \bibinfo{person}{Marcello Restelli}.} \bibinfo{year}{2020}\natexlab{}.
\newblock \showarticletitle{Importance Sampling Techniques for Policy
  Optimization}.
\newblock \bibinfo{journal}{\emph{Journal of Machine Learning Research}}
  \bibinfo{volume}{21}, \bibinfo{number}{141} (\bibinfo{year}{2020}),
  \bibinfo{pages}{1--75}.
\newblock
\urldef\tempurl%
\url{http://jmlr.org/papers/v21/20-124.html}
\showURL{%
\tempurl}


\bibitem[Mnih et~al\mbox{.}(2013)]%
        {DBLP:journals/corr/MnihKSGAWR13}
\bibfield{author}{\bibinfo{person}{Volodymyr Mnih}, \bibinfo{person}{Koray
  Kavukcuoglu}, \bibinfo{person}{David Silver}, \bibinfo{person}{Alex Graves},
  \bibinfo{person}{Ioannis Antonoglou}, \bibinfo{person}{Daan Wierstra}, {and}
  \bibinfo{person}{Martin~A. Riedmiller}.} \bibinfo{year}{2013}\natexlab{}.
\newblock \showarticletitle{Playing Atari with Deep Reinforcement Learning}.
\newblock \bibinfo{journal}{\emph{CoRR}}  \bibinfo{volume}{abs/1312.5602}
  (\bibinfo{year}{2013}).
\newblock
\showeprint[arXiv]{1312.5602}
\urldef\tempurl%
\url{http://arxiv.org/abs/1312.5602}
\showURL{%
\tempurl}


\bibitem[Naumov et~al\mbox{.}(2019)]%
        {naumov2019deep}
\bibfield{author}{\bibinfo{person}{Maxim Naumov}, \bibinfo{person}{Dheevatsa
  Mudigere}, \bibinfo{person}{Hao-Jun~Michael Shi}, \bibinfo{person}{Jianyu
  Huang}, \bibinfo{person}{Narayanan Sundaraman}, \bibinfo{person}{Jongsoo
  Park}, \bibinfo{person}{Xiaodong Wang}, \bibinfo{person}{Udit Gupta},
  \bibinfo{person}{Carole-Jean Wu}, \bibinfo{person}{Alisson~G Azzolini},
  {et~al\mbox{.}}} \bibinfo{year}{2019}\natexlab{}.
\newblock \showarticletitle{Deep learning recommendation model for
  personalization and recommendation systems}.
\newblock \bibinfo{journal}{\emph{arXiv preprint arXiv:1906.00091}}
  (\bibinfo{year}{2019}).
\newblock


\bibitem[Pham et~al\mbox{.}(2018)]%
        {pham2018efficient}
\bibfield{author}{\bibinfo{person}{Hieu Pham}, \bibinfo{person}{Melody Guan},
  \bibinfo{person}{Barret Zoph}, \bibinfo{person}{Quoc Le}, {and}
  \bibinfo{person}{Jeff Dean}.} \bibinfo{year}{2018}\natexlab{}.
\newblock \showarticletitle{Efficient neural architecture search via parameters
  sharing}. In \bibinfo{booktitle}{\emph{International conference on machine
  learning}}. PMLR, \bibinfo{pages}{4095--4104}.
\newblock


\bibitem[Real et~al\mbox{.}(2019)]%
        {real2019regularized}
\bibfield{author}{\bibinfo{person}{Esteban Real}, \bibinfo{person}{Alok
  Aggarwal}, \bibinfo{person}{Yanping Huang}, {and} \bibinfo{person}{Quoc~V
  Le}.} \bibinfo{year}{2019}\natexlab{}.
\newblock \showarticletitle{Regularized evolution for image classifier
  architecture search}. In \bibinfo{booktitle}{\emph{Proceedings of the aaai
  conference on artificial intelligence}}, Vol.~\bibinfo{volume}{33}.
  \bibinfo{pages}{4780--4789}.
\newblock


\bibitem[Schaul et~al\mbox{.}(2016)]%
        {DBLP:journals/corr/SchaulQAS15}
\bibfield{author}{\bibinfo{person}{Tom Schaul}, \bibinfo{person}{John Quan},
  \bibinfo{person}{Ioannis Antonoglou}, {and} \bibinfo{person}{David Silver}.}
  \bibinfo{year}{2016}\natexlab{}.
\newblock \showarticletitle{Prioritized Experience Replay}. In
  \bibinfo{booktitle}{\emph{4th International Conference on Learning
  Representations, {ICLR} 2016, San Juan, Puerto Rico, May 2-4, 2016,
  Conference Track Proceedings}}, \bibfield{editor}{\bibinfo{person}{Yoshua
  Bengio} {and} \bibinfo{person}{Yann LeCun}} (Eds.).
\newblock
\urldef\tempurl%
\url{http://arxiv.org/abs/1511.05952}
\showURL{%
\tempurl}


\bibitem[Schlegel et~al\mbox{.}(2019)]%
        {10.5555/3454287.3454448}
\bibfield{author}{\bibinfo{person}{Matthew Schlegel}, \bibinfo{person}{Wesley
  Chung}, \bibinfo{person}{Daniel Graves}, \bibinfo{person}{Jian Qian}, {and}
  \bibinfo{person}{Martha White}.} \bibinfo{year}{2019}\natexlab{}.
\newblock \bibinfo{booktitle}{\emph{Importance Resampling for Off-Policy
  Prediction}}.
\newblock \bibinfo{publisher}{Curran Associates Inc.}, \bibinfo{address}{Red
  Hook, NY, USA}.
\newblock


\bibitem[So et~al\mbox{.}(2019)]%
        {so2019evolved}
\bibfield{author}{\bibinfo{person}{David So}, \bibinfo{person}{Quoc Le}, {and}
  \bibinfo{person}{Chen Liang}.} \bibinfo{year}{2019}\natexlab{}.
\newblock \showarticletitle{The evolved transformer}. In
  \bibinfo{booktitle}{\emph{International conference on machine learning}}.
  PMLR, \bibinfo{pages}{5877--5886}.
\newblock


\bibitem[Song et~al\mbox{.}(2020)]%
        {song2020towards}
\bibfield{author}{\bibinfo{person}{Qingquan Song}, \bibinfo{person}{Dehua
  Cheng}, \bibinfo{person}{Hanning Zhou}, \bibinfo{person}{Jiyan Yang},
  \bibinfo{person}{Yuandong Tian}, {and} \bibinfo{person}{Xia Hu}.}
  \bibinfo{year}{2020}\natexlab{}.
\newblock \showarticletitle{Towards automated neural interaction discovery for
  click-through rate prediction}. In \bibinfo{booktitle}{\emph{Proceedings of
  the 26th ACM SIGKDD International Conference on Knowledge Discovery \& Data
  Mining}}. \bibinfo{pages}{945--955}.
\newblock


\bibitem[Tan and Le(2019)]%
        {tan2019efficientnet}
\bibfield{author}{\bibinfo{person}{Mingxing Tan} {and} \bibinfo{person}{Quoc
  Le}.} \bibinfo{year}{2019}\natexlab{}.
\newblock \showarticletitle{Efficientnet: Rethinking model scaling for
  convolutional neural networks}. In \bibinfo{booktitle}{\emph{International
  conference on machine learning}}. PMLR, \bibinfo{pages}{6105--6114}.
\newblock


\bibitem[Wen et~al\mbox{.}(2020)]%
        {wen2020neural}
\bibfield{author}{\bibinfo{person}{Wei Wen}, \bibinfo{person}{Hanxiao Liu},
  \bibinfo{person}{Yiran Chen}, \bibinfo{person}{Hai Li},
  \bibinfo{person}{Gabriel Bender}, {and} \bibinfo{person}{Pieter-Jan
  Kindermans}.} \bibinfo{year}{2020}\natexlab{}.
\newblock \showarticletitle{Neural predictor for neural architecture search}.
  In \bibinfo{booktitle}{\emph{European Conference on computer vision}}.
  Springer, \bibinfo{pages}{660--676}.
\newblock


\bibitem[White et~al\mbox{.}(2021)]%
        {white2021bananas}
\bibfield{author}{\bibinfo{person}{Colin White}, \bibinfo{person}{Willie
  Neiswanger}, {and} \bibinfo{person}{Yash Savani}.}
  \bibinfo{year}{2021}\natexlab{}.
\newblock \showarticletitle{Bananas: Bayesian optimization with neural
  architectures for neural architecture search}. In
  \bibinfo{booktitle}{\emph{Proceedings of the AAAI Conference on Artificial
  Intelligence}}, Vol.~\bibinfo{volume}{35}. \bibinfo{pages}{10293--10301}.
\newblock


\bibitem[Williams(1992)]%
        {10.1007/BF00992696}
\bibfield{author}{\bibinfo{person}{Ronald~J. Williams}.}
  \bibinfo{year}{1992}\natexlab{}.
\newblock \showarticletitle{Simple Statistical Gradient-Following Algorithms
  for Connectionist Reinforcement Learning}.
\newblock \bibinfo{journal}{\emph{Mach. Learn.}} \bibinfo{volume}{8},
  \bibinfo{number}{3–4} (\bibinfo{date}{may} \bibinfo{year}{1992}),
  \bibinfo{pages}{229–256}.
\newblock
\showISSN{0885-6125}
\urldef\tempurl%
\url{https://doi.org/10.1007/BF00992696}
\showDOI{\tempurl}


\bibitem[Wu et~al\mbox{.}(2019)]%
        {wu2019fbnet}
\bibfield{author}{\bibinfo{person}{Bichen Wu}, \bibinfo{person}{Xiaoliang Dai},
  \bibinfo{person}{Peizhao Zhang}, \bibinfo{person}{Yanghan Wang},
  \bibinfo{person}{Fei Sun}, \bibinfo{person}{Yiming Wu},
  \bibinfo{person}{Yuandong Tian}, \bibinfo{person}{Peter Vajda},
  \bibinfo{person}{Yangqing Jia}, {and} \bibinfo{person}{Kurt Keutzer}.}
  \bibinfo{year}{2019}\natexlab{}.
\newblock \showarticletitle{Fbnet: Hardware-aware efficient convnet design via
  differentiable neural architecture search}. In
  \bibinfo{booktitle}{\emph{Proceedings of the IEEE/CVF conference on computer
  vision and pattern recognition}}. \bibinfo{pages}{10734--10742}.
\newblock


\bibitem[Xie et~al\mbox{.}(2018)]%
        {xie2018snas}
\bibfield{author}{\bibinfo{person}{Sirui Xie}, \bibinfo{person}{Hehui Zheng},
  \bibinfo{person}{Chunxiao Liu}, {and} \bibinfo{person}{Liang Lin}.}
  \bibinfo{year}{2018}\natexlab{}.
\newblock \showarticletitle{SNAS: stochastic neural architecture search}.
\newblock \bibinfo{journal}{\emph{arXiv preprint arXiv:1812.09926}}
  (\bibinfo{year}{2018}).
\newblock


\bibitem[Ying et~al\mbox{.}(2019)]%
        {ying2019bench}
\bibfield{author}{\bibinfo{person}{Chris Ying}, \bibinfo{person}{Aaron Klein},
  \bibinfo{person}{Eric Christiansen}, \bibinfo{person}{Esteban Real},
  \bibinfo{person}{Kevin Murphy}, {and} \bibinfo{person}{Frank Hutter}.}
  \bibinfo{year}{2019}\natexlab{}.
\newblock \showarticletitle{Nas-bench-101: Towards reproducible neural
  architecture search}. In \bibinfo{booktitle}{\emph{International conference
  on machine learning}}. PMLR, \bibinfo{pages}{7105--7114}.
\newblock


\bibitem[Zhang et~al\mbox{.}(2022)]%
        {zhang2022dhen}
\bibfield{author}{\bibinfo{person}{Buyun Zhang}, \bibinfo{person}{Liang Luo},
  \bibinfo{person}{Xi Liu}, \bibinfo{person}{Jay Li}, \bibinfo{person}{Zeliang
  Chen}, \bibinfo{person}{Weilin Zhang}, \bibinfo{person}{Xiaohan Wei},
  \bibinfo{person}{Yuchen Hao}, \bibinfo{person}{Michael Tsang},
  \bibinfo{person}{Wenjun Wang}, {et~al\mbox{.}}}
  \bibinfo{year}{2022}\natexlab{}.
\newblock \showarticletitle{DHEN: A deep and hierarchical ensemble network for
  large-scale click-through rate prediction}.
\newblock \bibinfo{journal}{\emph{arXiv preprint arXiv:2203.11014}}
  (\bibinfo{year}{2022}).
\newblock


\bibitem[Zhang et~al\mbox{.}(2023a)]%
        {zhang2023nasrec}
\bibfield{author}{\bibinfo{person}{Tunhou Zhang}, \bibinfo{person}{Dehua
  Cheng}, \bibinfo{person}{Yuchen He}, \bibinfo{person}{Zhengxing Chen},
  \bibinfo{person}{Xiaoliang Dai}, \bibinfo{person}{Liang Xiong},
  \bibinfo{person}{Feng Yan}, \bibinfo{person}{Hai Li}, \bibinfo{person}{Yiran
  Chen}, {and} \bibinfo{person}{Wei Wen}.} \bibinfo{year}{2023}\natexlab{a}.
\newblock \showarticletitle{NASRec: weight sharing neural architecture search
  for recommender systems}. In \bibinfo{booktitle}{\emph{Proceedings of the ACM
  Web Conference 2023}}. \bibinfo{pages}{1199--1207}.
\newblock


\bibitem[Zhang et~al\mbox{.}(2023b)]%
        {zhang2023distdnas}
\bibfield{author}{\bibinfo{person}{Tunhou Zhang}, \bibinfo{person}{Wei Wen},
  \bibinfo{person}{Igor Fedorov}, \bibinfo{person}{Xi Liu},
  \bibinfo{person}{Buyun Zhang}, \bibinfo{person}{Fangqiu Han},
  \bibinfo{person}{Wen-Yen Chen}, \bibinfo{person}{Yiping Han},
  \bibinfo{person}{Feng Yan}, \bibinfo{person}{Hai Li}, {et~al\mbox{.}}}
  \bibinfo{year}{2023}\natexlab{b}.
\newblock \showarticletitle{DistDNAS: Search Efficient Feature Interactions
  within 2 Hours}.
\newblock \bibinfo{journal}{\emph{arXiv preprint arXiv:2311.00231}}
  (\bibinfo{year}{2023}).
\newblock


\bibitem[Zhaok et~al\mbox{.}(2021)]%
        {zhaok2021autoemb}
\bibfield{author}{\bibinfo{person}{Xiangyu Zhaok}, \bibinfo{person}{Haochen
  Liu}, \bibinfo{person}{Wenqi Fan}, \bibinfo{person}{Hui Liu},
  \bibinfo{person}{Jiliang Tang}, \bibinfo{person}{Chong Wang},
  \bibinfo{person}{Ming Chen}, \bibinfo{person}{Xudong Zheng},
  \bibinfo{person}{Xiaobing Liu}, {and} \bibinfo{person}{Xiwang Yang}.}
  \bibinfo{year}{2021}\natexlab{}.
\newblock \showarticletitle{Autoemb: Automated embedding dimensionality search
  in streaming recommendations}. In \bibinfo{booktitle}{\emph{2021 IEEE
  International Conference on Data Mining (ICDM)}}. IEEE,
  \bibinfo{pages}{896--905}.
\newblock


\bibitem[Zoph and Le(2016)]%
        {zoph2016neural}
\bibfield{author}{\bibinfo{person}{Barret Zoph} {and} \bibinfo{person}{Quoc~V
  Le}.} \bibinfo{year}{2016}\natexlab{}.
\newblock \showarticletitle{Neural architecture search with reinforcement
  learning}.
\newblock \bibinfo{journal}{\emph{arXiv preprint arXiv:1611.01578}}
  (\bibinfo{year}{2016}).
\newblock


\bibitem[Zoph et~al\mbox{.}(2018)]%
        {zoph2018learning}
\bibfield{author}{\bibinfo{person}{Barret Zoph}, \bibinfo{person}{Vijay
  Vasudevan}, \bibinfo{person}{Jonathon Shlens}, {and} \bibinfo{person}{Quoc~V
  Le}.} \bibinfo{year}{2018}\natexlab{}.
\newblock \showarticletitle{Learning transferable architectures for scalable
  image recognition}. In \bibinfo{booktitle}{\emph{Proceedings of the IEEE
  conference on computer vision and pattern recognition}}.
  \bibinfo{pages}{8697--8710}.
\newblock


\end{thebibliography}
%\bibliography{main}
% \clearpage
\appendix
\section*{Appendix}

\begin{figure}[b]
\centering
\includegraphics[width=0.47\textwidth]{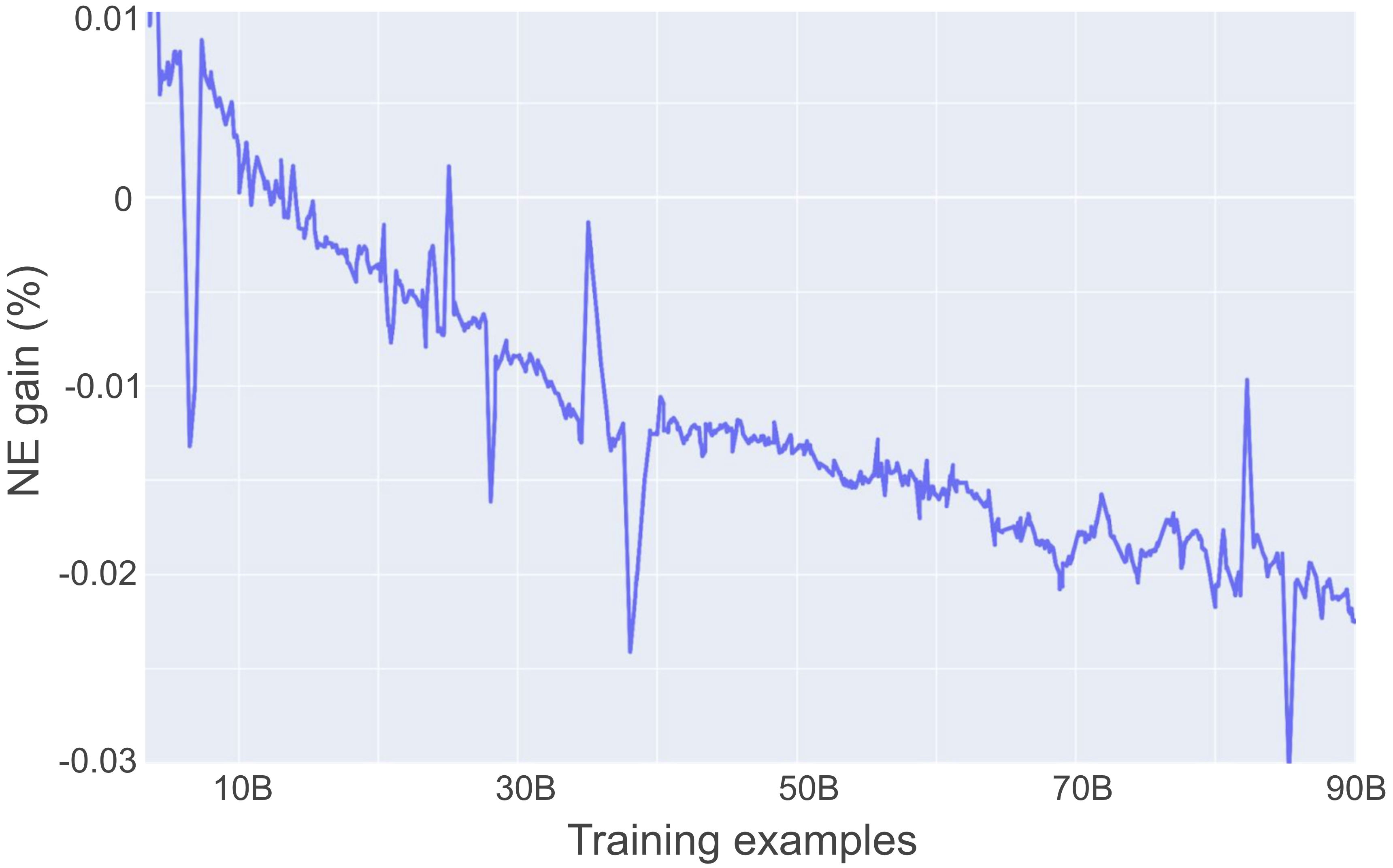}
\caption{Offline training NE-gain of ``model 1'' compared against a 31.3$\times$ model tuned by strong human.}
\label{fig:wsnas_vs_dhen31x}
\end{figure}

\subsection*{A. Ablation study of RL efficient search}

\begin{figure}[b]
\centering
\includegraphics[width=0.46\textwidth]{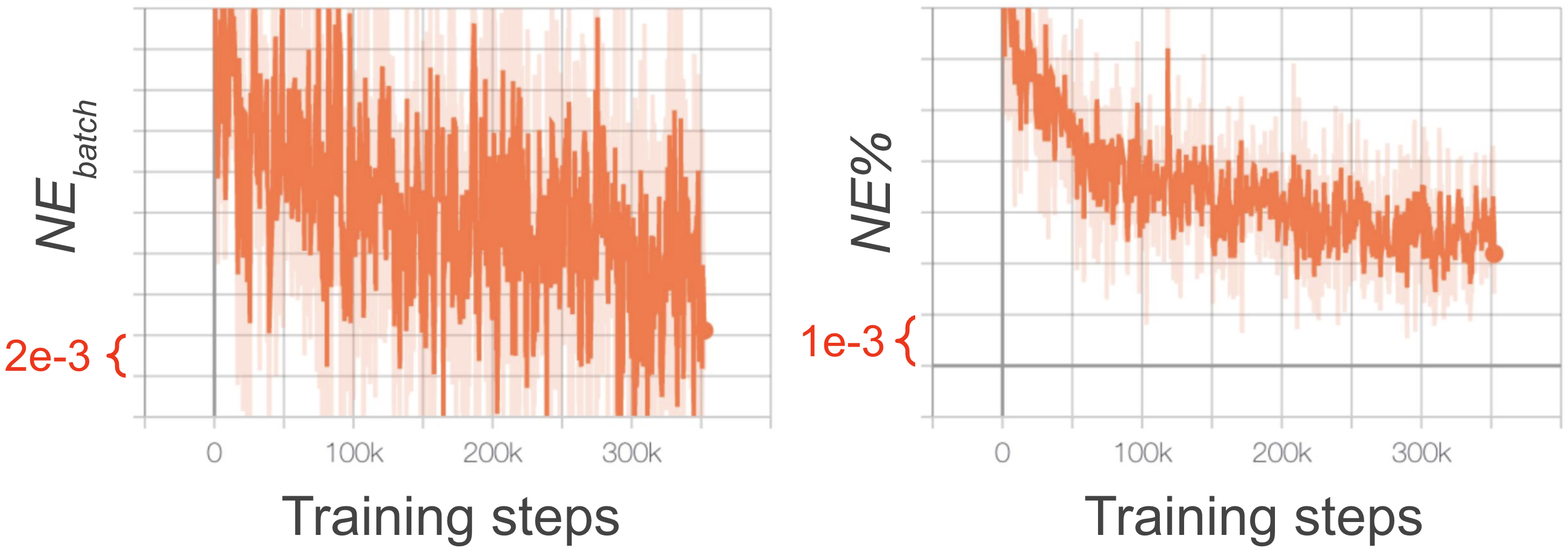}
\caption{Observed NE signal during supernet training (left: $NE_{batch}(a_{i})$, right: $NE_\%(a_{i})$)}
\label{fig:neBatch_vs_neGain_reward}
\end{figure}

\begin{figure}[b]
\centering
\includegraphics[width=0.46\textwidth]{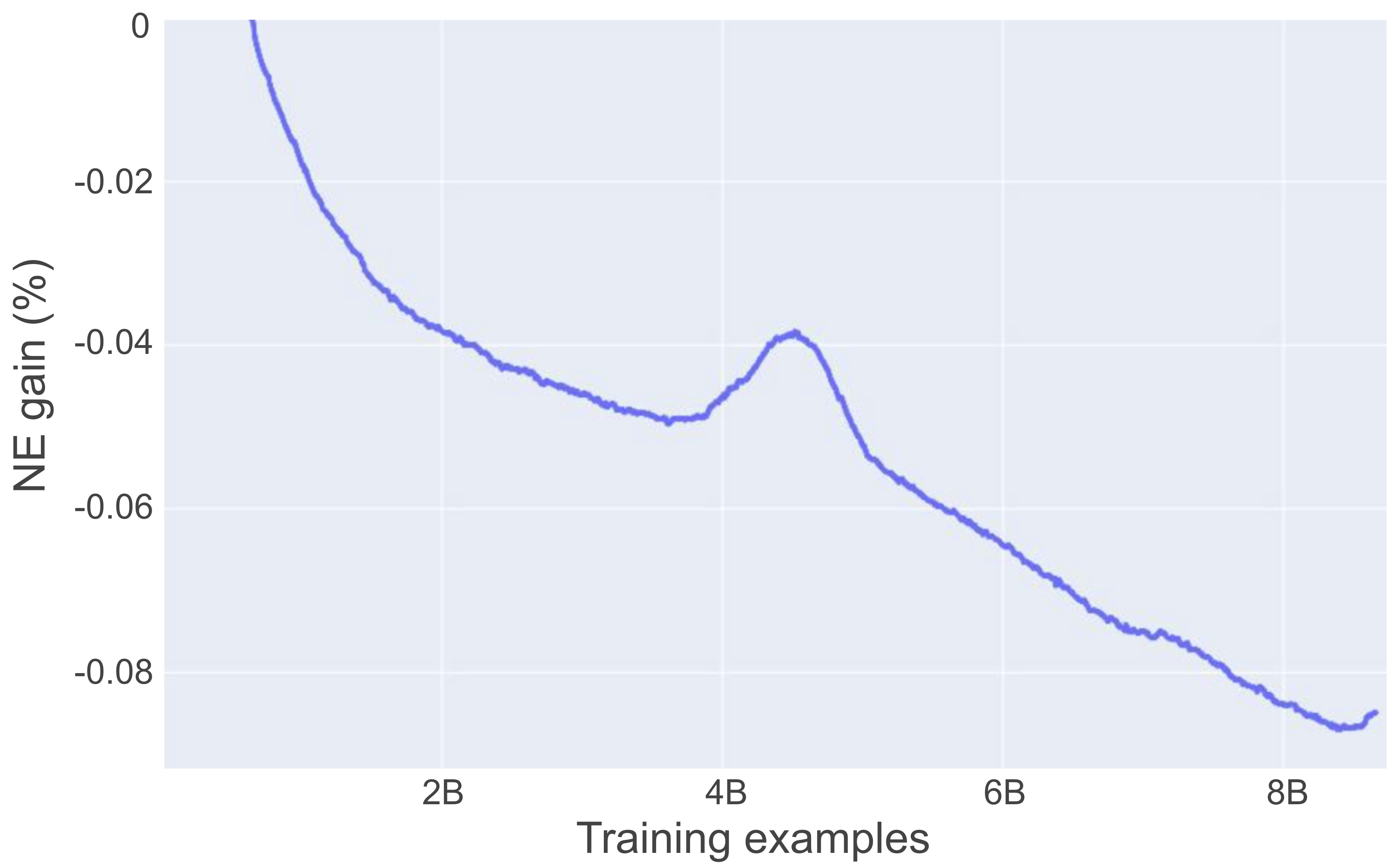}
\caption{Supernet NE gain of using $NE_\%(a_{i})$ and $NE_{batch}(a_{i})$ as reward }
\label{fig:neBatch_vs_neGain_ne}
\end{figure}

In Section~\ref{sec:one_shot_rl} we proposed $NE_\%(a_{i})$ to reduce noise by an in-place baseline in one-shot method and proposed an on/off-policy PG method to improve RL one-shot NAS sample efficiency. To validate the effectiveness of our proposals we present a detailed ablation study. First, we plot the observed $NE_{batch}(a_{i})$ and $NE_\%(a_{i})$ during a supernet training in Figure~\ref{fig:neBatch_vs_neGain_reward} where we can see that $NE_\%(a_{i})$ has significantly smaller variance compared to $NE_{batch}(a_{i})$.
Next, we verify that using the less noisy $NE_\%(a_{i})$ as RL rewards also lead to improved Rankitect RL search. Figure~\ref{fig:neBatch_vs_neGain_ne} shows the NE gain of supernet when using $NE_\%(a_{i})$ as RL reward versus using $NE_{batch}(a_{i})$; a negative value indicates that Rankitect with $NE_\%(a_{i})$ reward consistently samples models with better NE performance than when using $NE_{batch}(a_{i})$ as reward (note that Figure~\ref{fig:neBatch_vs_neGain_ne} only optimizes for NE, i.e., $\alpha=0$). 
Since computing $NE_\%(a_{i})$ (Eq.~\eqref{eq:ne_gain}) requires additional in-place training a baseline model, a natural question to ask is how does this affects Rankitect training/searching efficiency? 
%We report that in the same Fig.~\ref{fig:neBatch_vs_neGain_ne}, both Rankitect searches using $NE_\%(a_{i})$ and $NE_{batch}(a_{i})$ rewards achieved similar $\sim 1e5$ queries per second (QPS) throughout their search. 
We found that using $NE_\%(a_{i})$ as reward does not adversely affect search efficiency (compared to $NE_{batch}(a_{i})$); our speculation is that the training cost of the baseline model is negligible compared to the much larger supernet training cost.

\begin{figure}[b]
\centering
\includegraphics[width=0.46\textwidth]{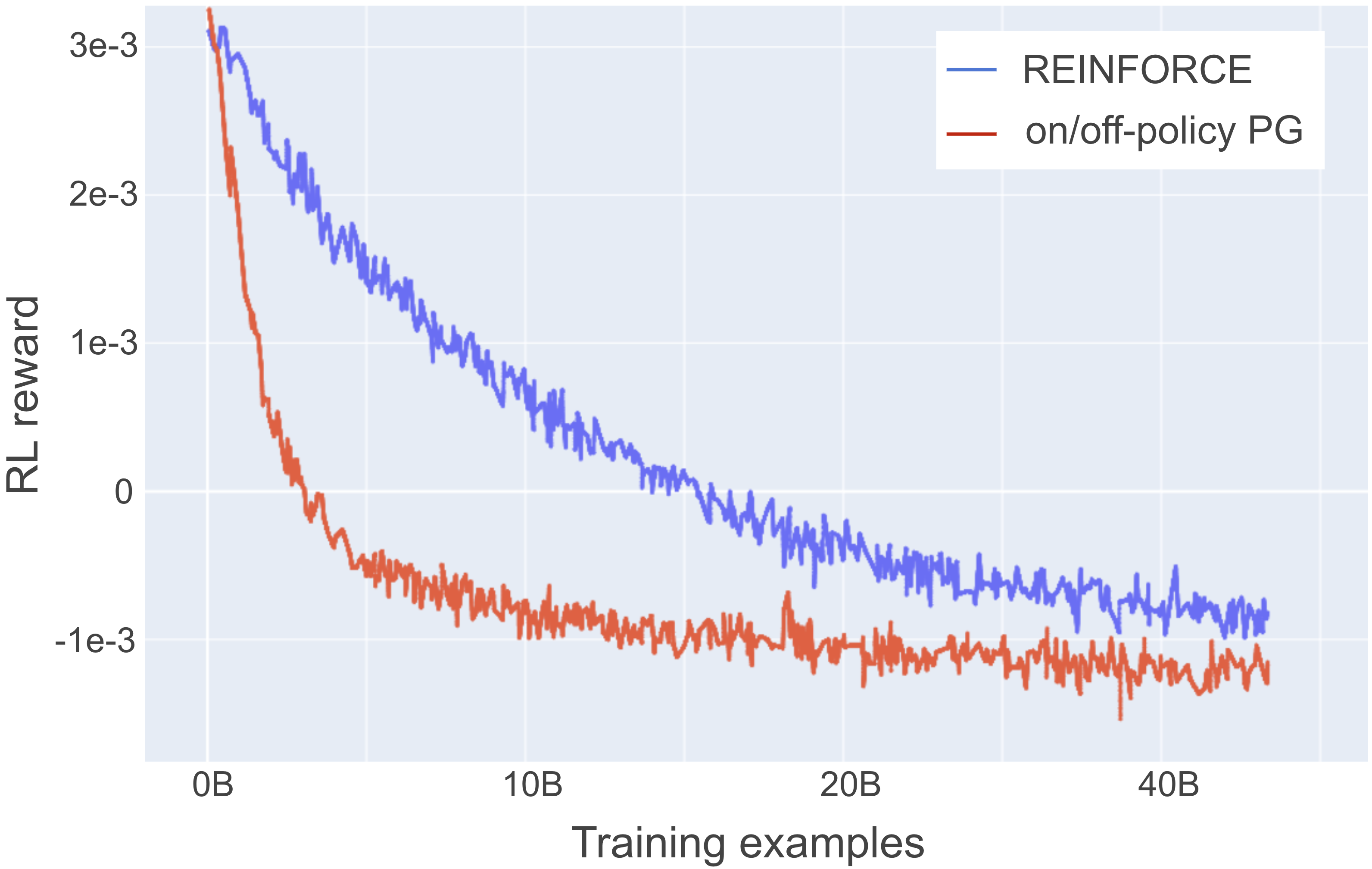}
\caption{RL training comparison (red: on/off-policy PG, blue: on-policy REINFORCE)}
\label{fig:pg_vs_offpg_reward}
\end{figure}

To validate the effectiveness of the proposed on/off-policy PG method, we plot the RL reward convergence of REINFORCE and on/off-policy PG in Figure~\ref{fig:pg_vs_offpg_reward}.
%where both methods achieved similar $\sim 1.7e5$ QPS.
From Figure~\ref{fig:pg_vs_offpg_reward} we can see that on/off-policy PG achieves faster learning . Moreover, we found similar search efficiency despite on/off-policy method receives 50$\times$ more policy updates than on-policy REINFORCE. Our speculation is that the computation cost of RL policy update is negligible compared to the much higher supernet training cost.

\end{document}